\documentclass[11pt,a4paper]{article}
\usepackage[hyperref]{acl2020}
\usepackage{times}
\usepackage{latexsym}
\usepackage{amsmath}
\usepackage{amssymb}
\usepackage{amsfonts}
\usepackage[mathscr]{euscript}
\usepackage{graphicx}
\usepackage[skip=5pt]{caption}
\usepackage{subcaption}
\usepackage{bm}
\usepackage{url}
\usepackage{xspace}
\usepackage{natbib}
\usepackage{tabularx}
\usepackage{booktabs}
\usepackage{multirow}
\usepackage{algorithm}
\usepackage{algpseudocode}
\usepackage{paralist}
\usepackage{eqparbox}
\newdimen{\algindent}
\usepackage{microtype}
\aclfinalcopy 
\def\aclpaperid{***}

\usepackage{tikz-dependency}
\usetikzlibrary{automata,decorations.markings,arrows,positioning,matrix,calc,patterns,angles,quotes,calc}
\newcommand{\argmax}[1]{\underset{#1}{\operatorname{arg}\,\operatorname{max}}\;}

\algnewcommand\LeftComment[2]{%
	\hspace{#1\algindent}$\triangleright$ \eqparbox{COMMENT}{#2} \hfill %
}

\newcommand{\secref}[1]{Section~\ref{sec:#1}}
\newcommand{\appref}[1]{Appendix~\ref{app:#1}}

\newcommand{\figref}[1]{Figure~\ref{fig:#1}}
\newcommand{\tabref}[1]{Table~\ref{tab:#1}}
\newcommand{\exref}[1]{Sentence~\ref{ex:#1}}

\definecolor{DarkGreen}{RGB}{0,111,0}
\definecolor{DarkBlue}{RGB}{0,0,111}
\definecolor{DarkRed}{RGB}{111,0,0}
\definecolor{DarkOrange}{RGB}{200,111,0}

\definecolor{lightblue}{rgb}{0.68, 0.85, 0.9}
\definecolor{lightred}{rgb}{0.89, 0.35, 0.13}
\definecolor{lightgreen}{rgb}{0.56, 0.93, 0.56}

\newcommand{\acl}[1]{Proc. of ACL\xspace}
\newcommand{\naacl}[1]{Proc. of NAACL\xspace}
\newcommand{\eccv}[1]{Proc. of ECCV\xspace}
\newcommand{\iccv}[1]{Proc. of ICCV\xspace}
\newcommand{\cvpr}[1]{Proc. of CVPR\xspace}
\newcommand{\emnlp}[1]{Proc. of EMNLP\xspace}
\newcommand{\iclr}[1]{Proc. of ICLR\xspace}
\newcommand{\icml}[1]{Proc. of ICML\xspace}
\newcommand{\jmlr}[1]{JMLR\xspace}
\newcommand{\nips}[1]{Proc. of NeurIPS\xspace}
\newcommand{\conll}[1]{Proc. of CoNLL\xspace}
\newcommand{\coling}[1]{Proc. of COLING\xspace}
\newcommand{\replnlp}[1]{Proc. of RepL4NLP\xspace}
\newcommand{\lsdsem}[1]{Proc. of LSDSem\xspace}
\newcommand{\blackbox}[1]{Proc. of BlackBoxNLP\xspace}
\newcommand{\emc}[1]{Proc. of EMC2\xspace}
\newcommand{\ranlp}[1]{Proc. of RANLP\xspace}
\newcommand{\repnlp}[1]{Proc. of Rep4NLP\xspace}
\newcommand{\recsys}[1]{Proc. of RecSys\xspace}
\newcommand{\nlposs}[1]{Proc. of NLP-OSS\xspace}
\newcommand{\starsem}[1]{Proc. of~$^*$SEM\xspace}

\newcommand{\iwp}[1]{Proc. of IWP\xspace}

\renewcommand{\phantom}[1]{{\textcolor{white}{#1}}}

\title{The Right Tool for the Job: Matching Model and Instance Complexities}

\author{Roy Schwartz$^{\diamondsuit\spadesuit}$ \quad
Gabriel Stanovsky$^{\diamondsuit\spadesuit}$ \quad
   Swabha Swayamdipta$^\diamondsuit$ \quad
Jesse Dodge$^\clubsuit$\thanks{~~Research completed during an internship at AI2.}  \quad
  Noah A. Smith$^{\diamondsuit\spadesuit}$ \\
  $^\diamondsuit$Allen Institute for Artificial Intelligence\\ 
  $^\spadesuit$Paul G. Allen School of Computer Science \& Engineering,
  University of Washington\\ 
  $^\clubsuit$School of Computer Science,
  Carnegie Mellon University\\ 
  {\tt \{roys,gabis,swabhas,jessed,noah\}@allenai.org}
}

\date{}

\begin{document}
\maketitle

\begin{abstract}
As NLP models become larger, executing a trained model requires significant computational resources incurring monetary and environmental costs. 
To better respect a given inference budget, we propose a modification to contextual representation fine-tuning which, during inference, allows for an early (and fast) ``exit'' from neural network calculations for simple instances, and late (and accurate) exit for hard instances.
To achieve this, we add classifiers to different layers of BERT and use their calibrated confidence scores to make early exit decisions.
We test our proposed modification on five different datasets in two tasks: three text classification datasets and two natural language inference benchmarks.
Our method presents a favorable speed/accuracy tradeoff in almost all cases, producing models which are up to five times faster than the state of the art, while preserving their accuracy. 
Our method also requires almost no additional training resources (in either time or parameters) compared to the baseline BERT model.
Finally, our method alleviates the need for costly retraining of multiple models at different levels of efficiency; 
we allow users to control the inference speed/accuracy tradeoff using a single trained model, by setting a single variable at inference time. We publicly release our code.\footnote{\href{https://github.com/allenai/sledgehammer}{github.com/allenai/sledgehammer}}

\end{abstract}

\section{Introduction}

\begin{figure}[!t]
\center
\newcommand{\figlen}[0]{\columnwidth}
\includegraphics[trim={2cm 3.3cm 3.4cm 3.3cm},clip,width=\figlen,page=2]{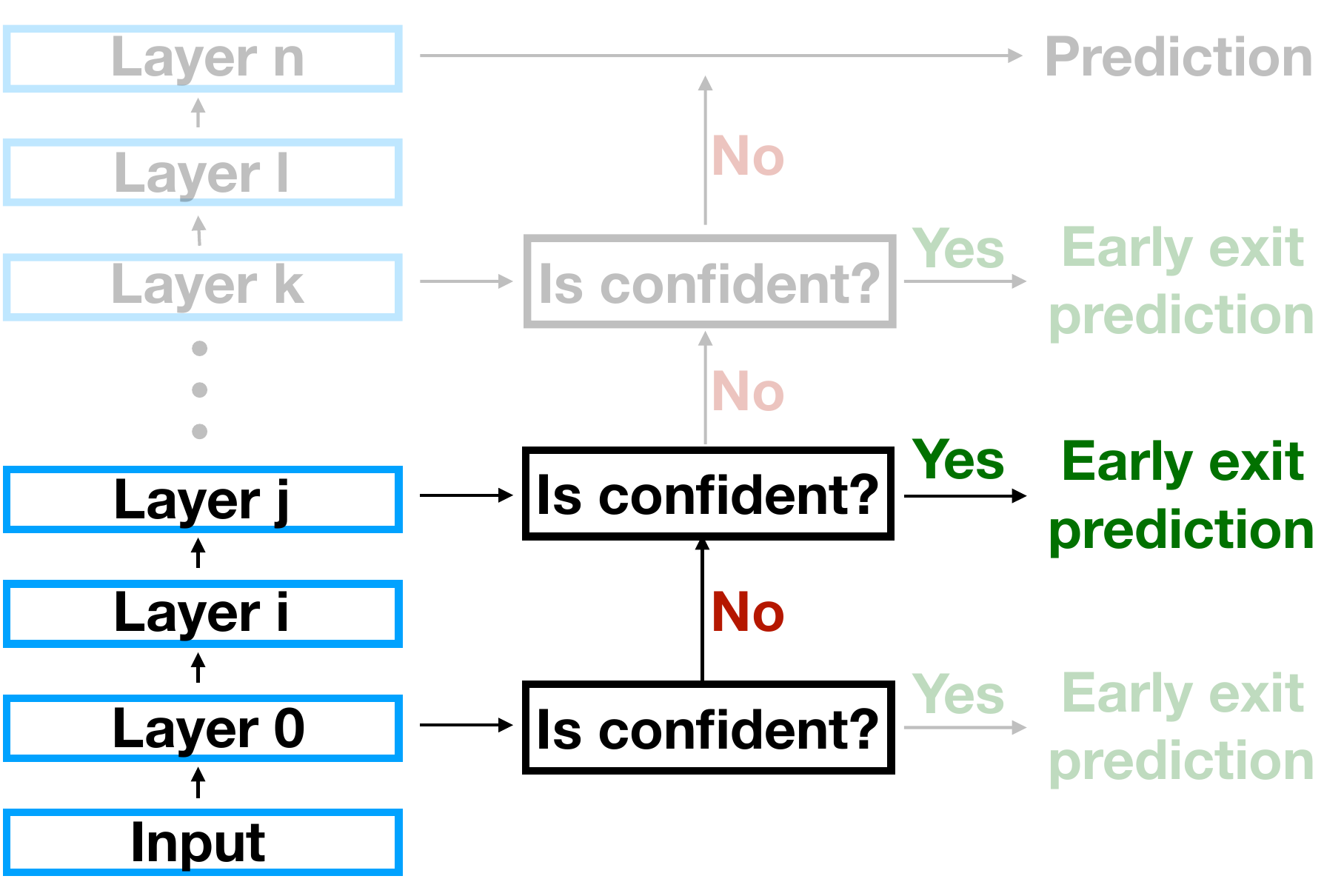}

\caption{
\label{fig:bert-illustration} 
An illustration of our approach.
Some layers of a BERT-large model are attached to output classifiers, which make their respective predictions.
The confidence of each layer-wise prediction is computed.
If high enough, the model takes an early exit, avoiding the computation associated with successive (higher) layers (grayed out).
Otherwise, the model continues to the next layer/classifier.
}
\end{figure}

The large increase in the size of artificial intelligence models often increases production costs \cite{Amodei:2018,Schwartz:2019}, and can also limit  adoption on real-time devices. 
Compared to \textit{training}, which is a one-time large investment, \textit{inference} costs are incurred for every instance in production, and can thus add up  significantly. 
For instance, Microsoft reports that using BERT \cite{Devlin:2019} to process Bing queries requires  more than 2,000 GPUs concurrently.\footnote{\url{https://tinyurl.com/tzhj3o8}}

We present a method to reduce the inference cost of today's common models in NLP:  fine-tuned contextual word representations. 
Our method exploits variation along two axes: \emph{models} differ in size and cost, and \emph{instances} vary in difficulty. 
Our method assesses the complexity of each test instance and matches it with the most efficient model in our ``toolbelt.''\footnote{Our approach should not be confused with model ensembles \cite{Kuncheva:2003}, where the prediction of multiple models is combined, \emph{on every instance}, in order to improve accuracy, at the expense of slower inference time.}
As a result, some instances, which we refer to in this paper as ``easy'' or ``simple,'' can be solved by small models, leading to  computational savings, while other instances (termed ``hard'' or ``difficult'') have access to larger models, thus retaining good performance.

We apply our method to the BERT-large model,
modifying its fine-tuning procedure by adding multiple output layers to some of its original $\ell=24$ layers.\footnote{For simplicity, we refer to these output layers as classifiers, though our method can also be applied to non-classification tasks.}
A classifier at the $k$th layer, is more efficient, though (presumably) less accurate than a classifier at a later $\ell$th layer (where $\ell > k$).
At inference time, we run each instance on these classifiers in increasing order of depth.
For each classification decision, we use its confidence as an inference-stopping criterion,
continuing to the next, larger classifier only if the current classifier is not confident enough in its prediction. Since confidence  scores play an important role, we use calibration techniques to make them more reliable.
Associating classifiers with  different layers of the same network allows us to reuse the computation performed by the simple classifiers for the complex ones.
See \figref{bert-illustration} for an illustration.

We experiment with three text classification benchmarks and two natural language inference (NLI) benchmarks.
We consider each of our classifiers with different BERT layers as individual baselines. 
We find that using our method leads to a consistently better speed/accuracy tradeoff in almost all cases.
In particular, in some cases, we obtain similar performance while being as much as five times faster than our strongest baseline (the original BERT-large mode with a single classification layer after the last layer).

Our approach, while allowing substantially faster inference compared to the standard BERT-large model, is neither slower to fine-tune nor significantly larger in terms of parameters, requiring less than 0.005\% additional parameters.
Moreover, our method is quite flexible: unlike other approaches for inference speed-up such as model distillation or pruning, which require training a different model for each point along the speed/accuracy curve, our method only requires training a single model, and by setting a single variable at inference time---the confidence threshold---supports each point along that curve.
Finally, our method is orthogonal to compression methods such as model distillation \cite{Hinton:2014}.
Our experiments with a distilled version of BERT \cite{Jiao:2019} show that our method further improves the speed/accuracy curve on top of that model. We publicly release our code.\footnote{\href{https://github.com/allenai/sledgehammer}{github.com/allenai/sledgehammer}}

\section{Premise: Models Vary in Size, Examples Vary in Complexity}

Our goal in this paper is to make model inference more efficient. 
Our premise relies on two general observations: first, as NLP models become bigger (e.g., in number of parameters), they become both better (in terms of downstream task accuracy), and slower to run. This trend is consistently observed, most notably in recent contextual representations work that compares different variants of the same model \cite[\textit{inter alia}]{Devlin:2019,Radford:2019,Raffel:2019}.

Second, inputs are not equally difficult.
For example,
instances differ in length and wealth of linguistic phenomena, which affects the amount of processing required to analyze them.
Consider the examples below for the task of sentiment analysis:
\begin{enumerate}[(1)]
\item \label{ex:simple} {The movie was awesome.}
\item \label{ex:complex} {I can't help but wonder whether the  plot was written by a 12 year-old or by an award-winning writer.}
\end{enumerate}

\exref{simple} is short and simple to process. In contrast, \exref{complex} is long, contains misleading positive phrases (``award-winning writer''), and uses figurative speech (``the plot was written by a 12 year-old''). As a result, it is potentially harder to process.\footnote{Note that simplicity is task-dependent.  For example, in topic classification, models often accumulate signal across a document, and shorter inputs (with less signal) may be more difficult than longer ones. See \secref{easy}.}

This work leverages these two observations by introducing a method to speed-up inference by matching simple instances with small models, and complex instances with large models. 

\section{Approach: The Right Tool for the Job}\label{sec:approach}

\paragraph{Motivation}
We assume a series of $n$ trained models {$m_1,\dots,m_n$} for a given task, such that for
each $1 < i \le n$, $m_i$ is both more accurate than $m_{i-1}$ (as
measured by a performance on validation data) and more expensive to execute.
Current practice in NLP, which favors accuracy rather than efficiency \cite{Schwartz:2019}, would typically run $m_n$ on each test instance, as it would likely lead to the highest test score.
However, many of the test instances could be solved by simpler (and faster) models;
if we had an oracle that identifies the smallest model that solves a given instance, we could use it to substantially speed up inference. 
Our goal is to create an automatic measure which approximates the behavior of such an oracle, and identify the cheapest accurate model for each instance. 

\paragraph{BERT-large}
To demonstrate our approach, we consider the BERT-large model \cite{Devlin:2019}, based on a transformer architecture \cite{Vaswani:2017} with 24 layers. 
To apply BERT-large to some downstream task, an output layer is typically added to the final layer of the model, and the model is fine-tuned on training data for that task. 
To make a prediction using the classifier on the final layer, the computation goes through all the layers sequentially, requiring more computation than a shallower model with fewer layers, which would suffice in some cases.

\paragraph{Suite of models}
Our approach leverages BERT's multilayered structure by adding an output layer to intermediate layers of the model. For $k < \ell$, the output layer after $k$ BERT layers exits the model earlier than a deeper output layer $\ell$, and therefore yields a more efficient (but potentially less accurate) prediction.

\paragraph{Confidence scores for early exit decisions}
To make early exit decisions, we calculate the layer-wise BERT representations sequentially.
As we reach a classification layer, we use it to make predictions.
We interpret the label scores output by softmax 
as  \textit{confidence scores}.
We use these confidence scores to decide whether to exit early or continue to the next (more expensive and more accurate) classifier.
See \figref{bert-illustration} for an illustration.

\paragraph{Training details}
To train the model, we use the standard way of applying BERT to downstream tasks---fine-tuning the pre-trained weights, while learning the weights of the randomly initialized classifier, where here we learn multiple classifiers instead of one.
As our loss function, we sum the losses of all classification layers, such that lower layers are trained to both be useful as feature generators for the higher layers, and as input to their respective classifiers.
This also means that every output layer is trained to perform well on all instances.
Importantly, we do not perform early exits during training, but only during inference.

To encourage monotonicity in performance of the different classifiers, each classifier at layer $k$  is given as input a weighted sum of all the layers up to and including $k$, such that the weight is learned during fine-tuning \cite{Peters:2018}.\footnote{We also considered feeding the output of previous classifiers as additional features to subsequent classifiers, known as stacking \citep{Wolpert92stackedgeneralization}. Preliminary experiments did not yield any benefits, so we did not further pursue this direction.}

\paragraph{Calibration}
Classifiers' confidence scores are not always reliable \cite{Jiang:2018}. One way to mitigate this concern is to use calibration, which encourages the confidence level to correspond to the probability that the model is correct  \cite{Degroot:1983}.
In this paper we use temperature calibration, which is a simple technique that has been shown to work well in practice \cite{Guo:2017}, in particular for BERT fine-tuning \cite{Desai:2020}.
The method learns a single parameter, denoted \textit{temperature} or $T$, and divides each of the logits $\{z_i\}$ by $T$ before applying the softmax function:
\[\mathrm{pred} = \argmax{i}\frac{\exp({z_i/T})}{\sum_j \exp({z_j/T})}\]
We select $T$ to maximize the log-likelihood of the development dataset.
Note that temperature calibration is monotonic and thus does not influence predictions. 
It is only used in our model to make early-exit decisions.

\paragraph{Discussion}
Our approach has several attractive properties. First, if $m_i$ is not sufficiently confident in its prediction, we reuse the computation and continue towards $m_{i+1}$ without recomputing the BERT layers up to $m_i$.
Second, while our model is larger in terms of parameters compared to the standard approach due to the additional classification layers, this difference is marginal compared to the total number of trainable parameters: our experiments used 4 linear output layers instead of 1, which results in an increase of 6K (binary classification) to 12K (4-way classification) parameters. For the BERT-large model with 335M trainable parameters, this is less than 0.005\% of the parameters. 
Third, as our experiments show (\secref{results}), while presenting a much better inference time/accuracy tradeoff, fine-tuning our model is as fast as fine-tuning the standard model with a single output layer. Moreover, our model allows for controlling this tradeoff by setting the confidence threshold at inference time,
allowing users to better utilize the model for their inference budget.

\section{Experiments}\label{sec:experiments}

\begin{table}[t]
\centering
\begin{tabular}{l c c c c}
\toprule
{\bf Name} & {\bf \#labels} & {\bf Train} & {\bf Val.} & {\bf Test} \\
\midrule
AG & 4 & 115K & \phantom{0.}5K & 7.6K \\ 
IMDB & 2 & \phantom{0}20K & \phantom{0.}5K & \phantom{.}25K \\ 
SST & 2 &  \phantom{00}7K & 0.9K & 1.8K \\
\midrule
SNLI & 3 & 550K & \phantom{.}10K &\phantom{.}10K \\
MNLI & 3 & 393K & 9.8K & 9.8K \\
\bottomrule
\end{tabular}
\caption{\label{tab:Datasets}
Number of labels and instances for the datasets in our experiments. The top set are text classification datasets, and the bottom set are NLI datasets.
}
\end{table}

\begin{figure*}[!t]
\newcommand{\figlen}[0]{.55\columnwidth}
\newcommand{\figheight}[0]{4.2cm}
\begin{subfigure}[t]{\figlen}
\centering
\fbox{\includegraphics[trim={2cm 3.3cm 17.4cm 3cm},clip,page=5,height=\figheight]{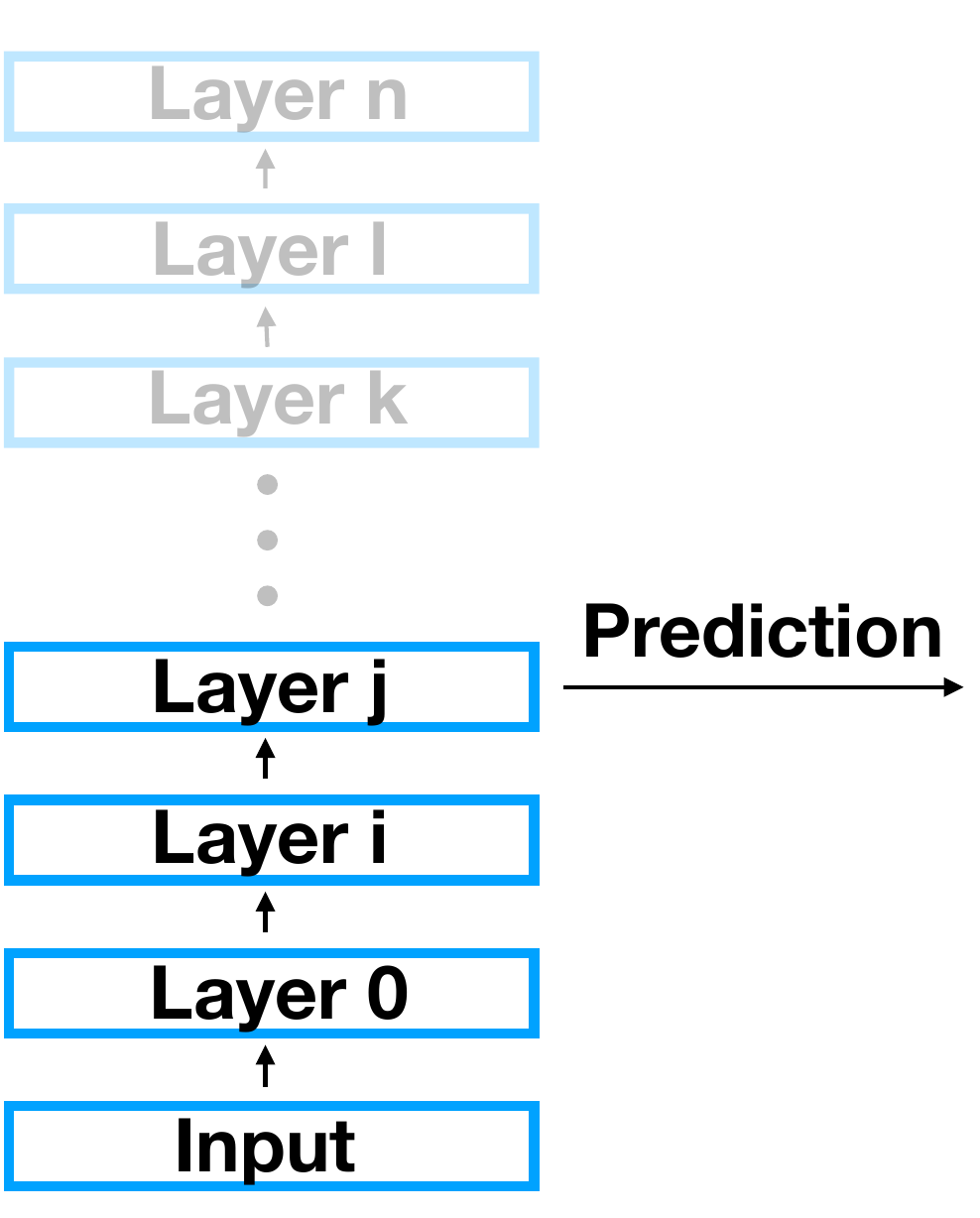}}
\caption{\label{fig:partial-baseline} Efficient Baseline}
\end{subfigure}
\quad
\begin{subfigure}[t]{\figlen}
\centering
\fbox{\includegraphics[trim={2cm 3.3cm 17.4cm 3cm},clip,page=6,height=\figheight]{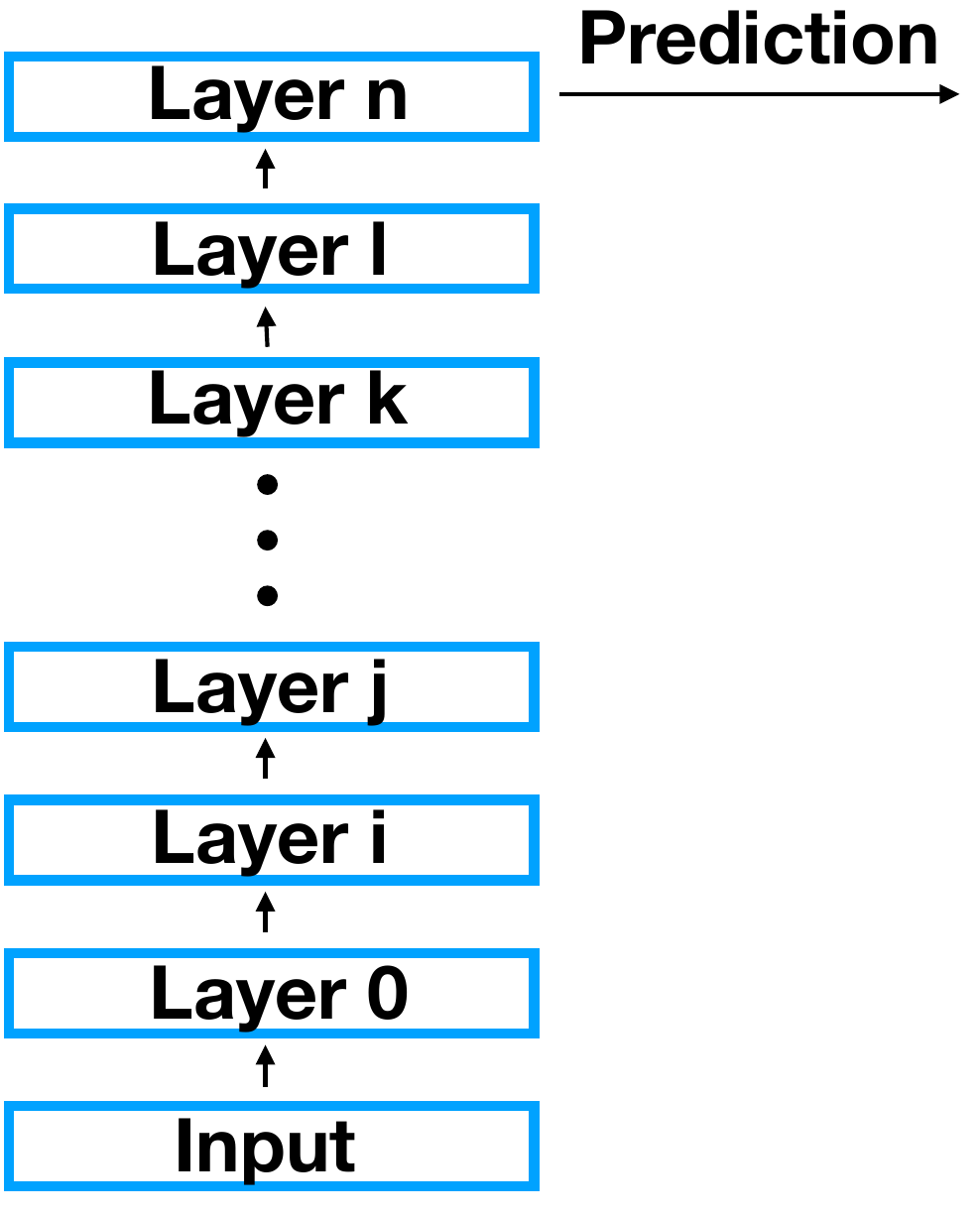}}
\caption{\label{fig:standard-baseline} Standard Baseline}
\end{subfigure}
\quad
\begin{subfigure}[t]{\figlen}
\centering
\fbox{\includegraphics[trim={2cm 3.3cm 3.4cm 3.3cm},clip,height=\figheight,page=7]{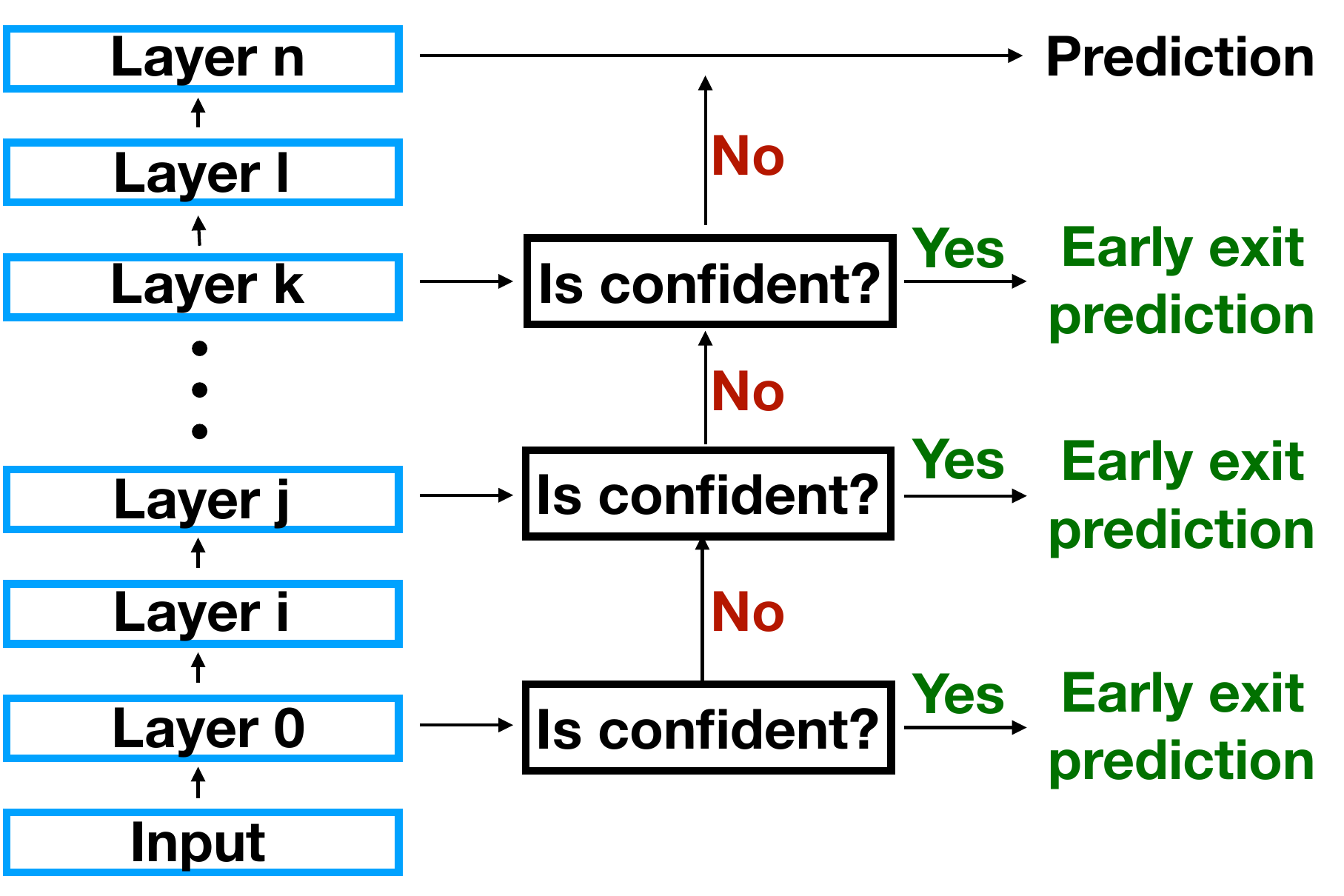}}
\caption{\label{fig:our-approach} Our approach}
\end{subfigure}

\caption{
\label{fig:baseline-illustration} 
Illustration of our baselines. 
(\ref{fig:partial-baseline}) Efficient baseline: adding a single output layer to an intermediate layer, while not processing the remaining BERT layers.
(\ref{fig:standard-baseline}) The standard model: adding a single output layer to the final BERT layer.
(\ref{fig:our-approach}) Our approach: adding multiple output layers to intermediate BERT layers; running the corresponding classifiers sequentially, while taking early exits based on their confidence scores.}
\end{figure*}

To test our approach, we experiment with three text classification and two natural language inference (NLI) tasks in English.
NLI is a pairwise sentence classification task, where the goal is to predict whether a hypothesis sentence entails, contradicts or is neutral to a premise sentence \cite{Dagan:05}.
Below we describe our datasets, our baselines, and our experimental setup.

\paragraph{Datasets}
For text classification, we experiment with the AG news topic identification dataset \cite{Zhang:2015} and two sentiment analysis datasets: IMDB \cite{Maas:2011} and the binary Stanford sentiment treebank (SST; \citealp{Socher:2013}).\footnote{For SST, we only used full sentences, not phrases.}
For NLI, we experiment with the SNLI \cite{Bowman:2015} and MultiNLI (MNLI; \citealp{Williams:2018}) datasets.
We use the standard train-development-test splits for all datasets except for MNLI, for which there is no public test set. As MNLI contains two validation sets (matched and mismatched), we use the matched validation set as our validation set and the mismatched validation set as our test set. See \tabref{Datasets} for dataset statistics.

\paragraph{Baselines}
We use two types of baselines: running BERT-large in the standard way, with a single output layer on top of the last layer, and three efficient baselines of increasing size (\figref{baseline-illustration}). 
Each is a fine-tuned BERT model with a single output layer after some intermediate layer.
Importantly, these baselines offer a speed/accuracy tradeoff, but not within a single model like our approach.

As all baselines have a single output layer, they all have a single loss term, such that BERT layers $1,\dots,k$ only focus on a single classification layer, rather than multiple ones as in our approach. 
As with our model, the single output layer in each of our baselines is given as input a learned weighted sum of all BERT layers up to the current layer.

As an upper bound to our approach, we consider a variant of our model that uses the \emph{exact} amount of computation required to solve a given instance. It does so by replacing the confidence-based early-exit decision function with an oracle that returns the fastest classifier that is able to solve that instance, or the fastest classifier for instances that are not correctly solved by any of the classifiers.

\paragraph{Experimental setup}
We experiment with BERT-large-uncased (24 layers).
We add output layers to four layers: 0, 4, 12 and 23.\footnote{Preliminary experiments with other configurations, including ones with more layers, led to similar results.} 
We use the first three layer indices for our efficient baselines (the last one corresponds to our standard baseline). 
See \appref{implementation} for implementation details.

For training, we use the largest batch size that fits in our GPU memory for each dataset, for both our baselines and our model.
Our approach relies on discrete early-exit decisions that might differ between instances in a batch.
For the sake of simplicity, we use a batch size of 1 during inference.
This is useful for production setups where instances arrive one by one.
Larger batch sizes can be applied using methods such as budgeted batch classification \cite{Huang:2018}, 
which specify a budget for the batch and select a subset of the instances to fit that budget, while performing early exit for the rest of the instances. 
We defer the technical implementation of this idea to future work.

To measure efficiency, we compute the average runtime of a single instance, across the test set. 
We repeat each validation and test experiment five times and report the mean and standard deviation.

At prediction time, our method takes as an input a threshold between 0 and 1, which is applied to each confidence score to decide whether to exit early.  Lower thresholds result in earlier exits, with 0 implying the most efficient classifier is always used.  A threshold of 1 always uses the most expensive and accurate classifier.
\section{Results}\label{sec:results}

\paragraph{A better speed/accuracy tradeoff.}

\begin{figure}[!ht]
\newcommand{\figlen}[0]{.87\columnwidth}
    \centering
    \includegraphics[trim={0cm 1cm 0cm 1cm}, clip,width=\figlen]{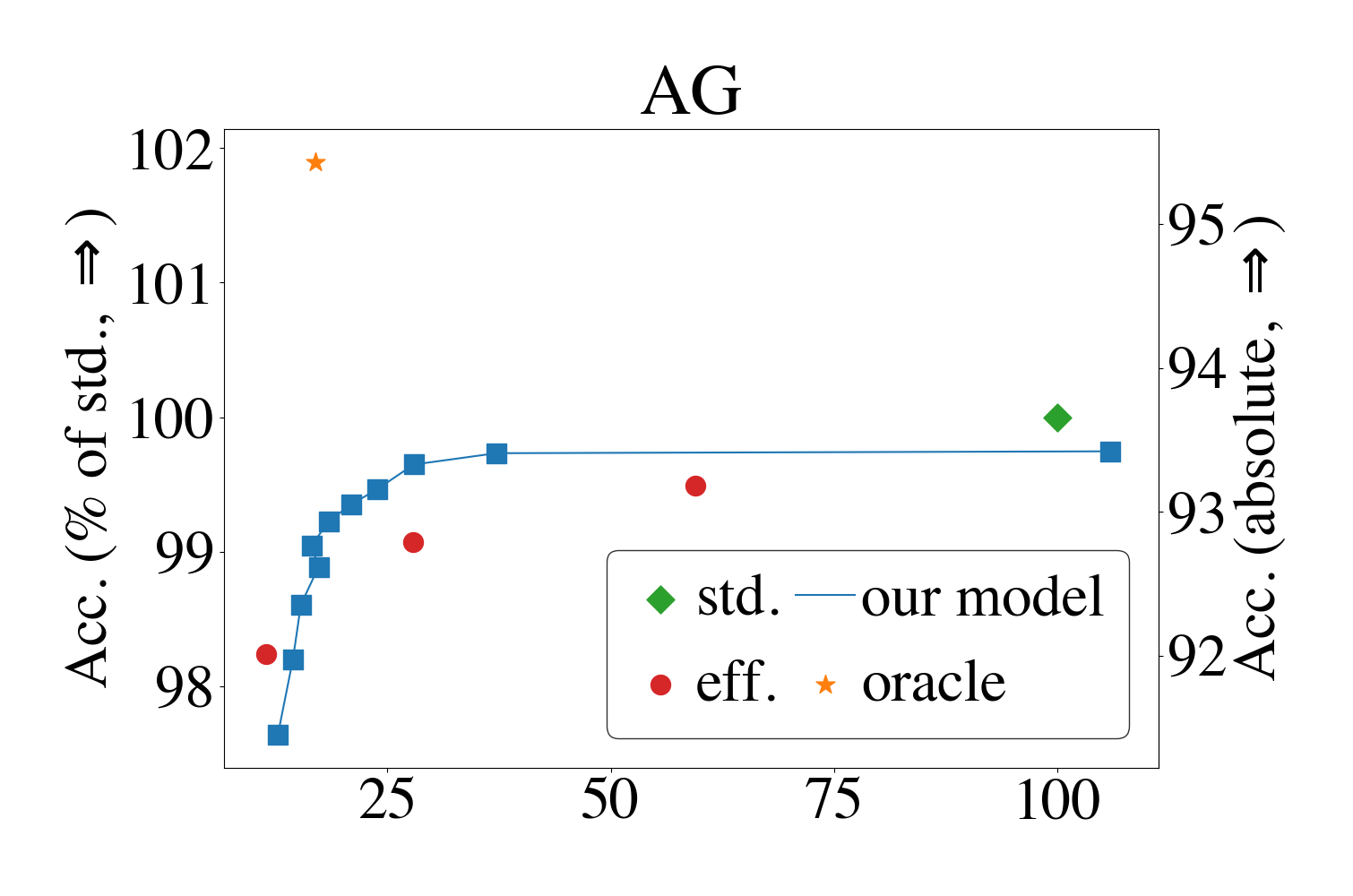}
    \includegraphics[trim={0cm 1cm 0cm 1cm}, clip,width=\figlen]{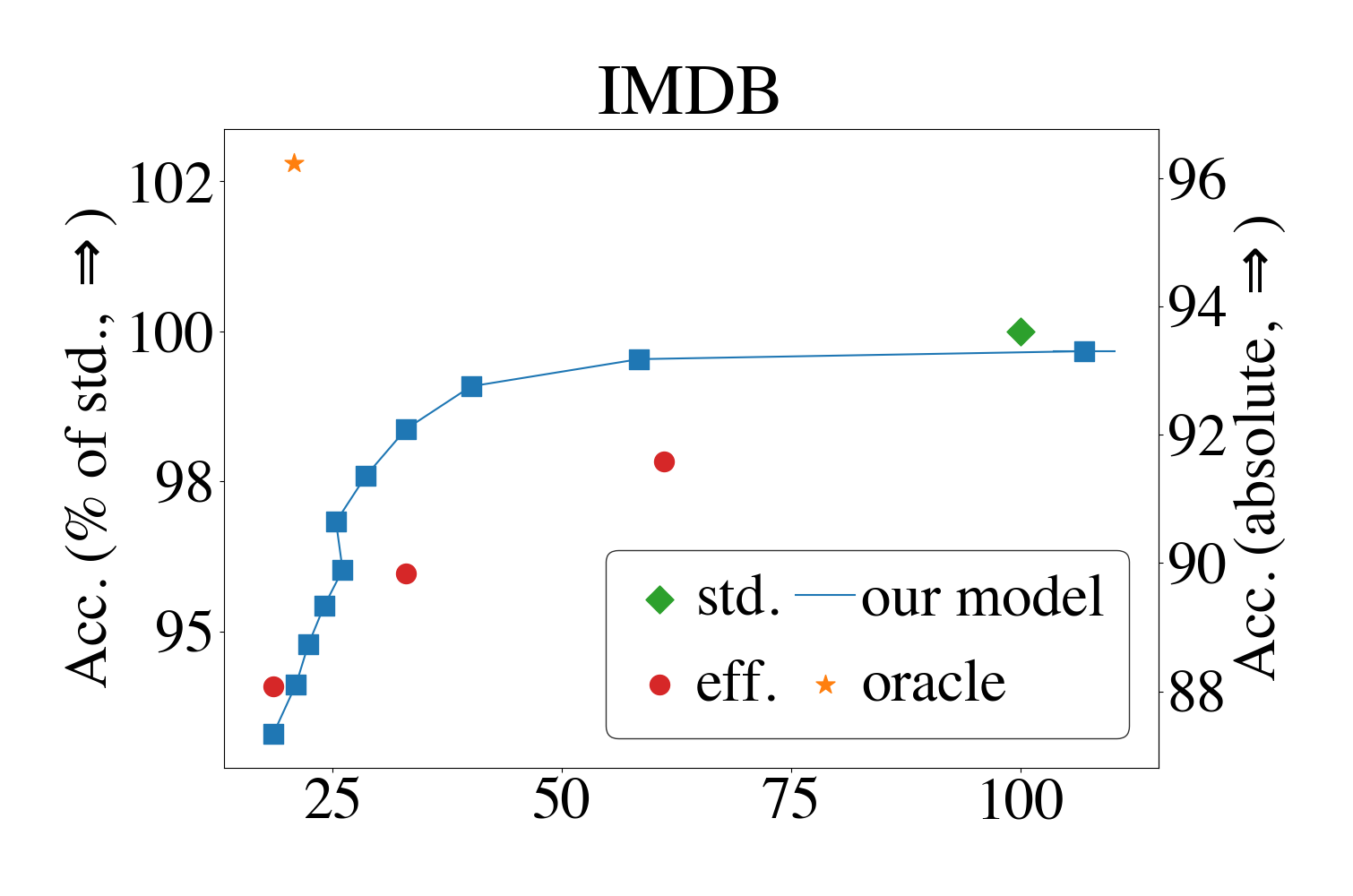}
    \includegraphics[trim={0cm 1cm 0cm 1cm}, clip,width=\figlen]{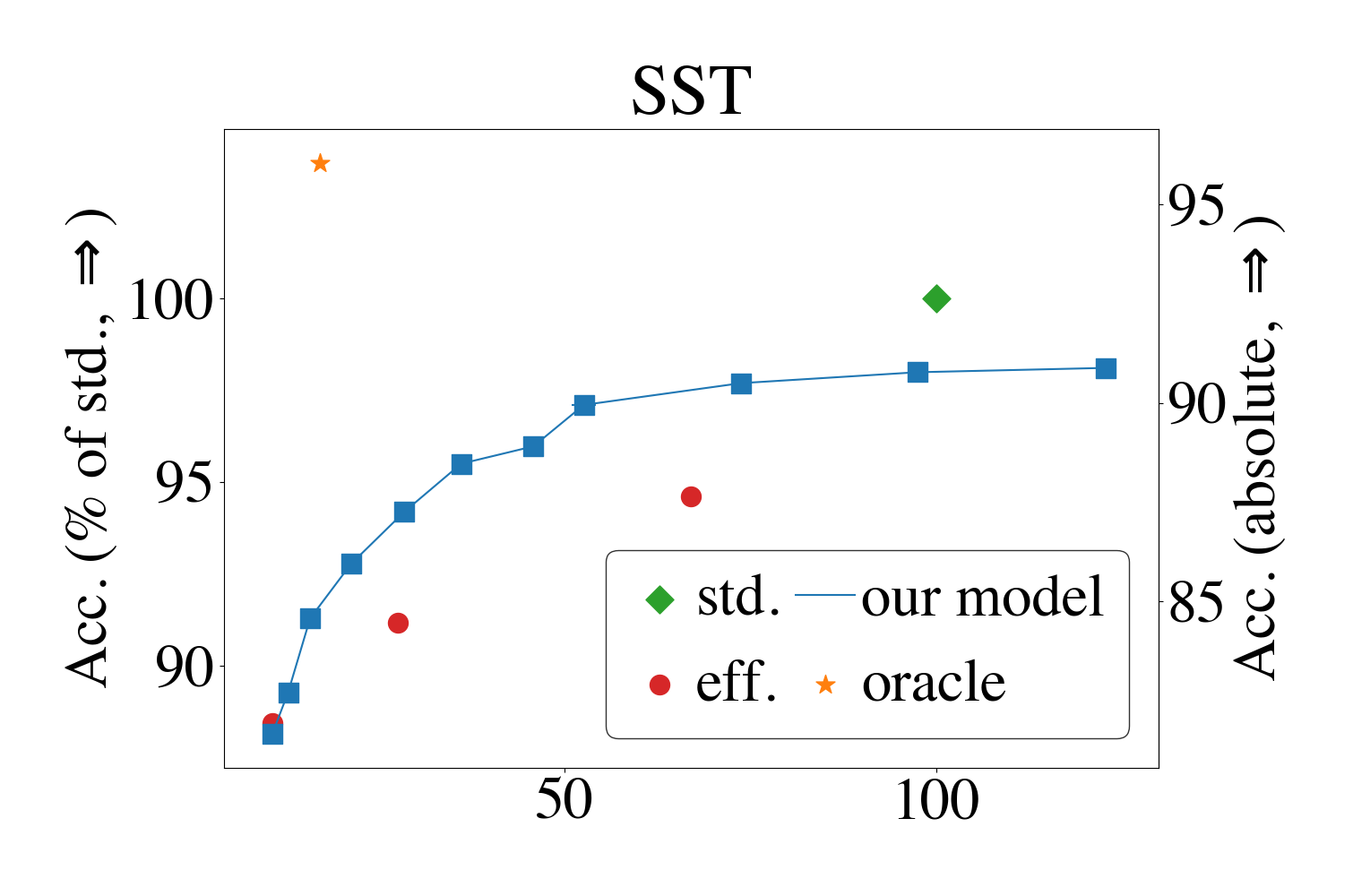}
    \includegraphics[trim={0cm 1cm 0cm 1cm}, clip,width=\figlen]{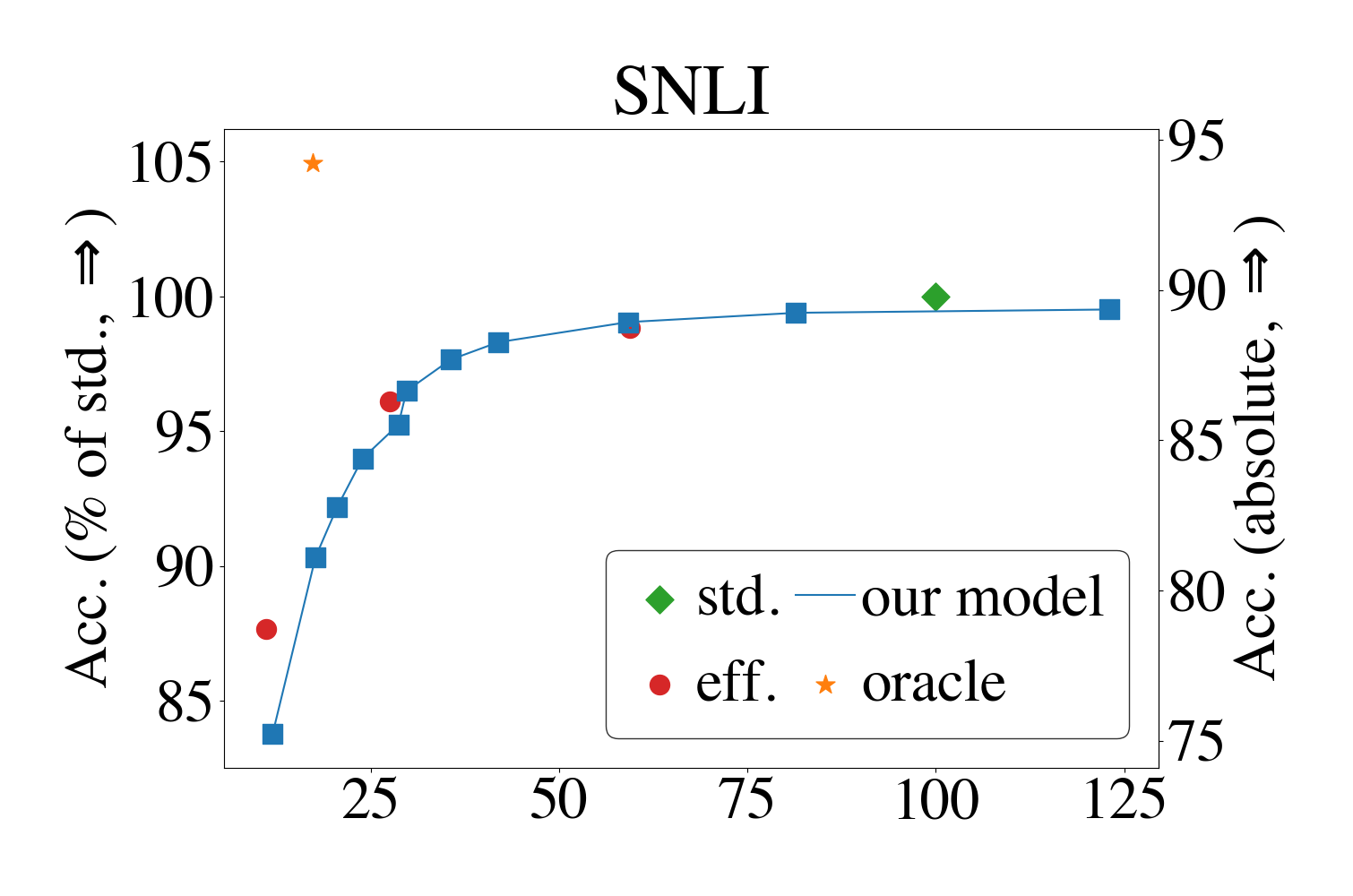}
    \includegraphics[trim={0cm 0 0cm 1cm}, clip,width=\figlen]{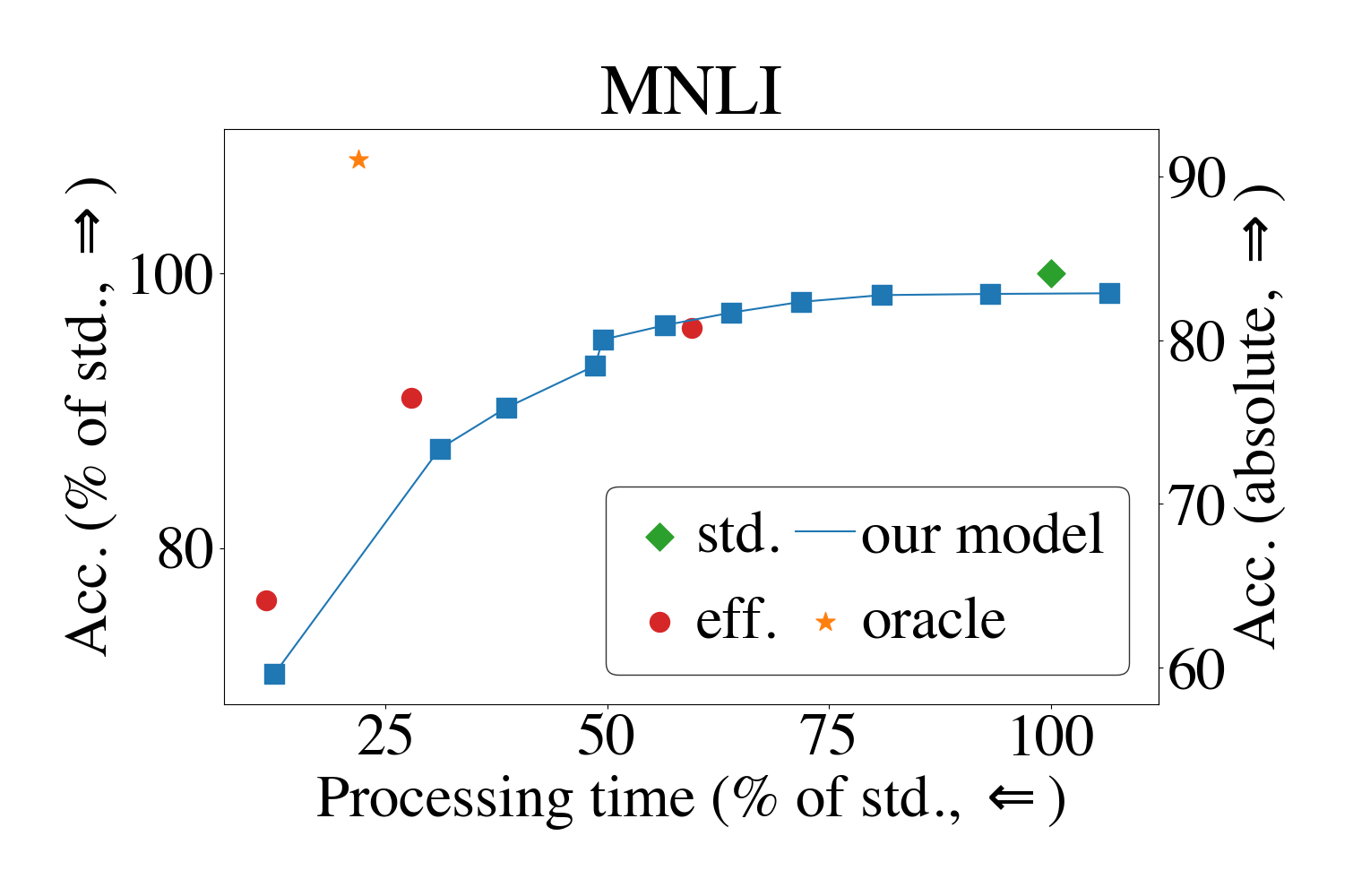}
    \caption{\label{fig:test_results}
        Test accuracy and processing time of our approach ({\color{blue}{blue squares}}, each point representing a different confidence threshold), our standard baseline (std., {\color{DarkGreen}{green diamond}}), efficient baselines (eff., {\color{red}{red dots}}), and  oracle baseline ({\color{orange}{orange star}}).
    Left and higher is better. Our method presents similar or better speed/accuracy tradeoff in almost all cases.
}
\end{figure}

\figref{test_results} presents our test results.\footnote{For increased reproduciblity \cite{Dodge:2019a}, we also report validation results in \appref{dev_results}.}
The blue line shows our model, where each point corresponds to an increasingly large confidence threshold. The leftmost (rightmost) point is threshold 0 (1), with $x$-value showing the fraction of processing time relative to the standard baseline.

Our first observation is that our efficient baselines constitute a fast alternative to the standard BERT-large model. 
On AG, a classifier trained on layer 12 of BERT-large is 40\% faster and within 0.5\% of the standard model. On SNLI and IMDB a similar speedup results in 2\% loss in performance.

Most notably, our approach presents a similar or better tradeoff in almost all cases. Our model is within 0.5\% of the standard model while being 40\% (IMDB) and 80\% (AG) faster.
For SST, our curve is strictly above two of the efficient baselines, while being below the standard one. In the two NLI datasets, our curve is slightly above the curve for the medium budgets, and below it for lower ones. 

Finally, the results of the oracle baseline indicate the further potential of our approach: in all cases, the oracle outperforms the original baseline by 1.8\% (AG) to 6.9\% (MNLI), while being 4--6 times faster.
These results motivate further exploration of better early-exit criteria (see \secref{easy}).
They also highlight the diversity of the different classifiers. One might expect that the set of correct predictions by the smaller classifiers will be contained in the corresponding sets of the larger classifiers. The large differences between the original baseline and our oracle indicate that this is not the case, and motivate future research on efficient ensemble methods which reuse much of the computation across different models.

\paragraph{Extreme case analysis}
Our results hint that combining the loss terms of each of our classifiers hurts their performance compared to our baselines, which use a single loss term.
For the leftmost point in our graphs---always selecting the most efficient classifier---we observe a substantial drop in performance compared to the corresponding most efficient baseline, especially for the NLI datasets. For our rightmost point (always selecting the most accurate classifier), we observe a smaller drop, mostly in SST and MNLI, compared to the corresponding baseline, but also slower runtime, probably due to the overhead of running the earlier classifiers.  

These trends further highlight the potential of our method, which is able to outperform the baseline speed-accuracy curves despite the weaker starting point. It also suggests ways to further improve our method by studying more sophisticated methods to combine the loss functions of our classifiers, and encourage them to be as precise as our baselines. We defer this to future work.

\paragraph{Similar training time}

Fine-tuning BERT-large with our approach has a similar cost to fine-tuning the standard BERT-large model, with a single output layer. 
\tabref{training_time} shows the fine-tuning time of our model and the standard BERT-large baseline.
Our model is not slower to fine-tune in four out of five cases, and is even slightly faster in three of them.\footnote{We note that computing the calibration temperature requires additional time, which ranges between 3 minutes (SST) to 24 minutes (MNLI).}

\begin{table}[tb!]
\centering
\begin{tabular}{l c c}
\toprule
\multirow{2}{*}{{\bf Dataset}} & \multicolumn{2}{c}{{\bf Training Time}} \\
& {Ours} & {Standard}\\
\midrule
AG & \phantom{0}52 & \phantom{0}53 \\
IMDB & \phantom{0}56 & \phantom{0}57 \\
SST & \phantom{00}4 & \phantom{00}4 \\
SNLI & 289 &  300 \\
MNLI & 852   & 835 \\
\bottomrule
\end{tabular}
\caption{\label{tab:training_time}
Fine-tuning times (in minutes) of our model compared to the most accurate baseline: the standard BERT-large model with a single output layer. }
\end{table}

This property makes our approach appealing compared to other approaches for reducing runtime such as pruning or model distillation (\secref{related}). These require, in addition to training the full model, also training another model for each point along the speed/accuracy curve, therefore substantially increasing the overall training time required to generate a full speed/accuracy tradeoff. In contrast, our single model allows for full control over this tradeoff by adjusting the confidence threshold, 
without increasing the training time compared to the standard, most accurate model.

\paragraph{Combination with model distillation}
A key property of our approach is that it can be applied to any multi-layer model.
Particularly, it can be combined with other methods for making models more efficient, such as model distillation. 
To demonstrate this, we repeat our IMDB experiments with tinyBERT \cite{Jiao:2019}, which is a distilled version of BERT-base.\footnote{While we experimented with BERT-large and not BERT-base, the point of this experiment is to illustrate the potential of our method to be combined with distillation, and not to directly compare to our main results.} We experiment with the tinyBERT v2 6-layer-768dim version.\footnote{\citet{Jiao:2019} also suggested a task-specific version of tinyBERT which distills the model based on the downstream task. For consistency with our BERT-large experiments, we use the general version.}

\begin{figure}[!ht]
\newcommand{\figlen}[0]{\columnwidth}
    \centering
    \includegraphics[trim={0cm -0.5cm 0cm 1cm}, clip,width=\figlen]{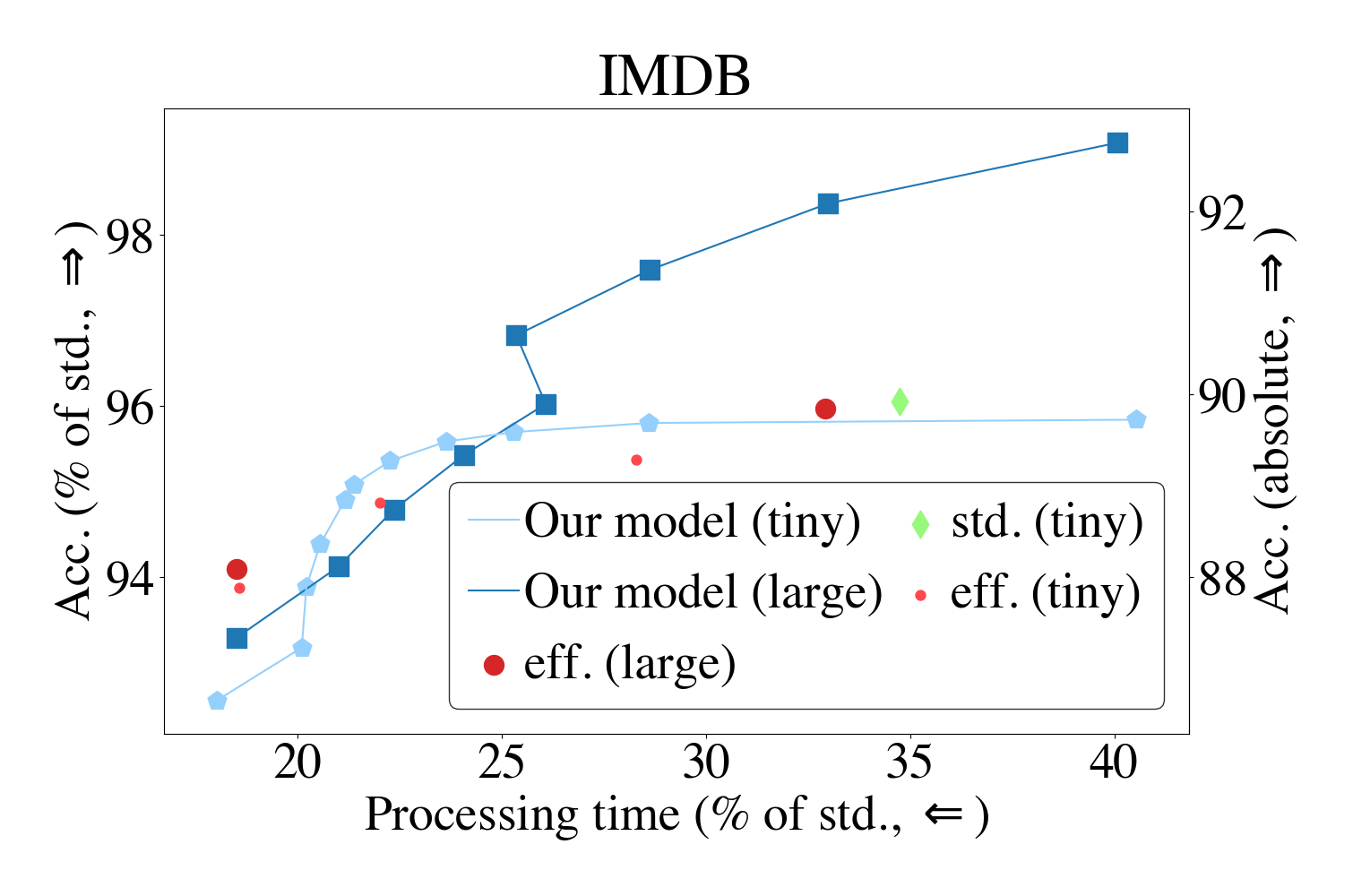}
    \caption{\label{fig:distilled-results} 
    Experiments with tinyBERT. Our method (\textcolor{lightblue}{light-blue pentagons}) provides a better speed-accuracy tradeoff compared to the standard (\textcolor{lightgreen}{light-green diamonds}) and efficient (\textcolor{lightred}{small light-red dots}) baselines. 
    For comparison, we also show the results of our method (\textcolor{blue}{blue squares}) and our  efficient baselines (\textcolor{red}{large red dots}) with BERT-large. Our method applied to BERT-large provides the overall best tradeoff. 
}
\end{figure}

\figref{distilled-results} shows our IMDB results. 
Much like for BERT-large, our method works well for tinyBERT, providing a better speed/accuracy tradeoff compared to the standard tinyBERT baseline and the efficient tinyBERT baselines.

Second, while tinyBERT is a distilled version of BERT-\emph{base}, its speed-accuracy tradeoff is remarkably similar to our BERT-\emph{large} efficient baselines, 
which hints that our efficient baselines are a simpler alternative to tinyBERT, and as effective for model compression.
Finally, our method applied to BERT-large provides the best overall speed-accuracy tradeoff, especially with higher budgets.


\begin{table}[!t]
\centering
\begin{tabular}{l c c c }
\toprule
{\bf Dataset} & {\bf Length} & {\bf Consistency} \\
\midrule
AG & \phantom{--}0.13 & 0.37\\
IMDB & --0.17 & 0.47\\
SST & --0.19 & 0.36\\
SNLI & --0.08 & 0.44\\
MNLI & --0.13 & 0.39\\
\bottomrule
\end{tabular}
\caption{\label{tab:correlation_analysis}
Spearman's $\rho$ correlation between confidence levels for our most efficient classifier and two measures of difficulty: document length and consistency. 
Confidence is correlated reasonably with consistency across all datasets.
For all datasets except AG, confidence is (loosely) negatively correlated with document length. For the AG topic classification dataset, confidence is (loosely) positively correlated.
Results for the other layers show a similar trend.}
\end{table}

\section{A Criterion for ``Difficulty''}\label{sec:easy}

Our approach is motivated by the inherent variance in the level of complexity of text instances, and leverages this variance to obtain a better speed/accuracy tradeoff compared to our baselines.
Our method also automatically identifies instances on which smaller models are highly confident in their predictions.
Here we analyze our data using other definitions of difficulty.
Perhaps surprisingly, we find that the various definitions are not strongly correlated with ours.
The results we observe below, combined with the performance of our oracle baseline (\secref{results}), motivate further study on more advanced methods for early exiting, which could potentially yield even larger computational gains.

\paragraph{Shorter is easier?}

We first consider the length of instances: is our model more confident in its decisions on short documents compared to longer ones?
To address this we compute Spearman's $\rho$ correlation between the confidence level of our most efficient classifier and the document's length. 

The results in \tabref{correlation_analysis} show that the correlations across all datasets are generally low ($|\rho| < 0.2$). Moreover, as expected, across four out of five datasets, the (weak) correlation between confidence and length is negative; our model is somewhat more confident in its prediction on shorter documents. 
The fifth dataset (AG), shows the opposite trend: confidence is positively correlated with length. 
This discrepancy might be explained by the nature of the tasks we consider. For instance, IMDB and SST are sentiment analysis datasets, where longer texts might include conflicting evidence and thus be harder to classify. In contrast, AG is a news topic detection dataset, where a conflict between topics is uncommon, and longer documents provide more opportunities to find the topic.

\paragraph{Consistency and difficulty}
Our next criterion for ``difficulty'' is the consistency of model predictions. 
\citet{Toneva:2019} proposed a notion of ``unforgettable'' training instances, which once the model has predicted correctly, it never predicts incorrectly for the remainder of training iterations. 
Such instances can be thought of as ``easy'' or memorable examples. 
Similarly, \citet{Sakaguchi:2019} defined test instances as ``predictable'' if multiple simple models predict them correctly.  
Inspired by these works, we define the criterion of consistency: whether all classifiers in our model agree on the prediction of a given instance, regardless of whether it is correct or not. 
\tabref{correlation_analysis} shows  Spearman's $\rho$ correlation between the confidence of the most efficient classifier and this measure of consistency.
Our analysis reveals a medium correlation between confidence and consistency across all datasets ($0.37 \le \rho \le 0.47$),
which indicates that the measure of confidence generally agrees with the measure of consistency.

\paragraph{Comparison with hypothesis-only criteria}
\citet{Gururangan:2018} and \citet{Poliak:2018} showed that some NLI instances can be solved by only looking at the hypothesis---these were artifacts of the annotation process. 
They argued that such instances are ``easier'' for machines, compared to those which required access to the full input, which they considered ``harder.''
\tabref{snli_correlation_analysis} shows the correlation between the confidence of each of our classifiers on the SNLI and MNLI dataset with the confidence of a hypothesis-only classifier.
Similarly to the consistency results, we see that the confidence of our most efficient classifier is reasonably correlated with the predictions of the hypothesis-only classifier. 
As expected, as we move to larger, more accurate classifiers, which presumably are able to make successful predictions on harder instances, this correlation decreases.

\paragraph{Inter-annotator consensus}
Both NLI datasets include labels from five different annotators.
We treat the inter-annotator consensus (IAC) as another measure of difficulty: the higher the consensus is, the easier the instance. 
We compute IAC for each example as the fraction of annotators who agreed on the majority label, hence this number ranges from 0.6 to 1.0 for five annotators.
\tabref{snli_correlation_analysis} shows the correlation between the confidence of our classifiers with the IAC measure on SNLI and MNLI.  
The correlation with our most efficient classifiers is rather weak, only 0.08 and 0.14. Surprisingly, as we move to larger models, the correlation increases, up to 0.32 for the most accurate classifiers. 
This indicates that the two measures perhaps capture a different notion of difficulty.

\begin{table}[!t]
\setlength{\tabcolsep}{6pt}

\centering
\begin{tabular}{l c c c c}
\toprule
\multirow{2}{.8cm}{{\bf Layer}} & \multicolumn{2}{c}{{\bf SNLI}} &
\multicolumn{2}{c}{{\bf MNLI}} \\
& {Hyp.-Only} & {IAC} & {Hyp.-Only} & {IAC}\\
\midrule
0 & 0.39 & 0.14 & 0.37 & 0.08 \\
4 & 0.31 & 0.25 & 0.35 & 0.21\\
12 & 0.31 & 0.31& 0.32 & 0.27\\
23 & 0.28 & 0.32 & 0.30 & 0.32\\
\bottomrule
\end{tabular}
\caption{\label{tab:snli_correlation_analysis}
Spearman's $\rho$ correlation between confidence levels for our classifiers (of different layers) on the validation sets of SNLI and MNLI, and two measures of difficulty: hypothesis-only classifier predictions (Hyp.-Only) and inter-annotator consensus (IAC).}
\end{table}

\begin{figure*}[!ht]
\newcommand{\figlen}[0]{.4\columnwidth}
    \centering
    \includegraphics[trim={0cm 0 0cm 0}, clip,width=\figlen]{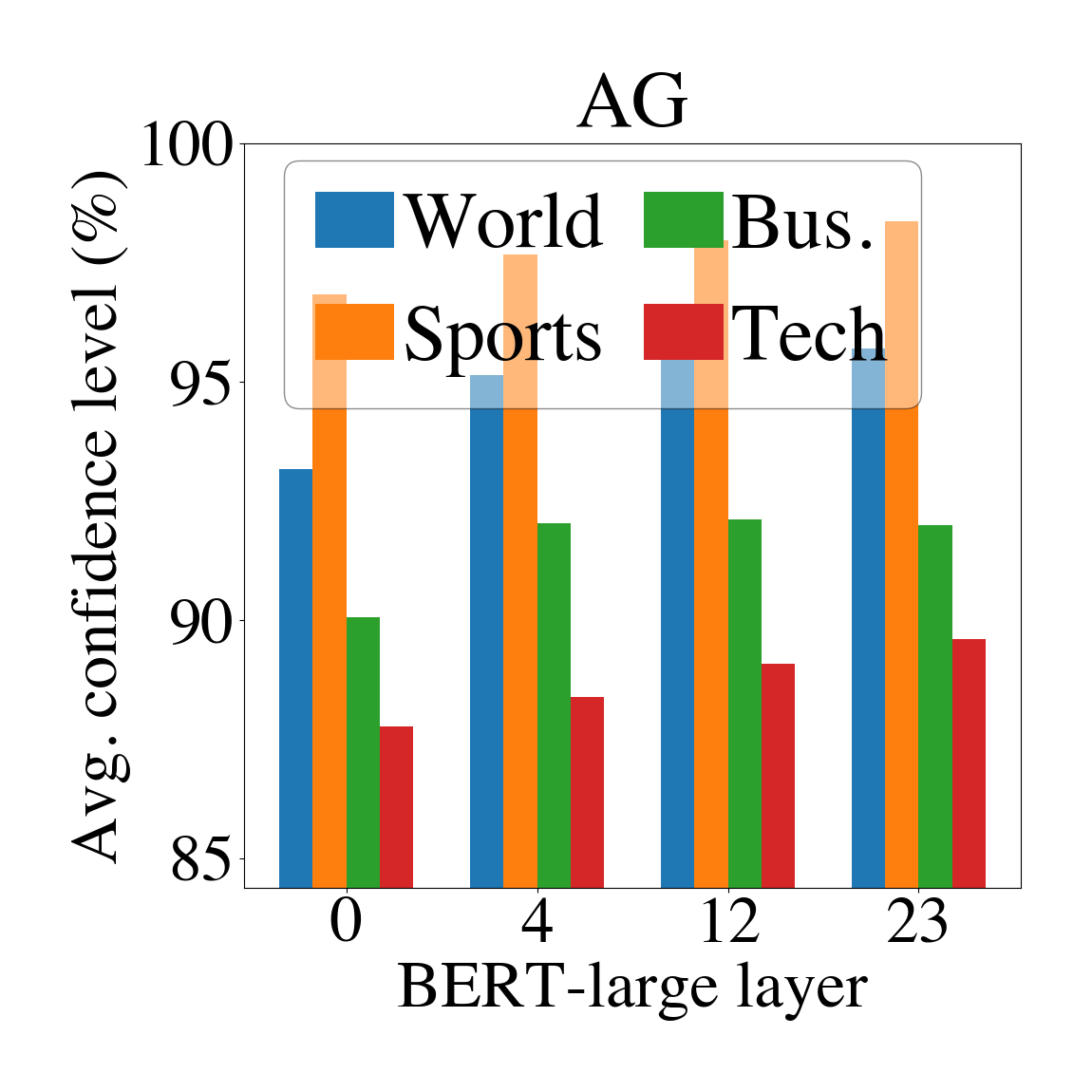}
    \includegraphics[trim={0cm 0 0cm 0}, clip,width=\figlen]{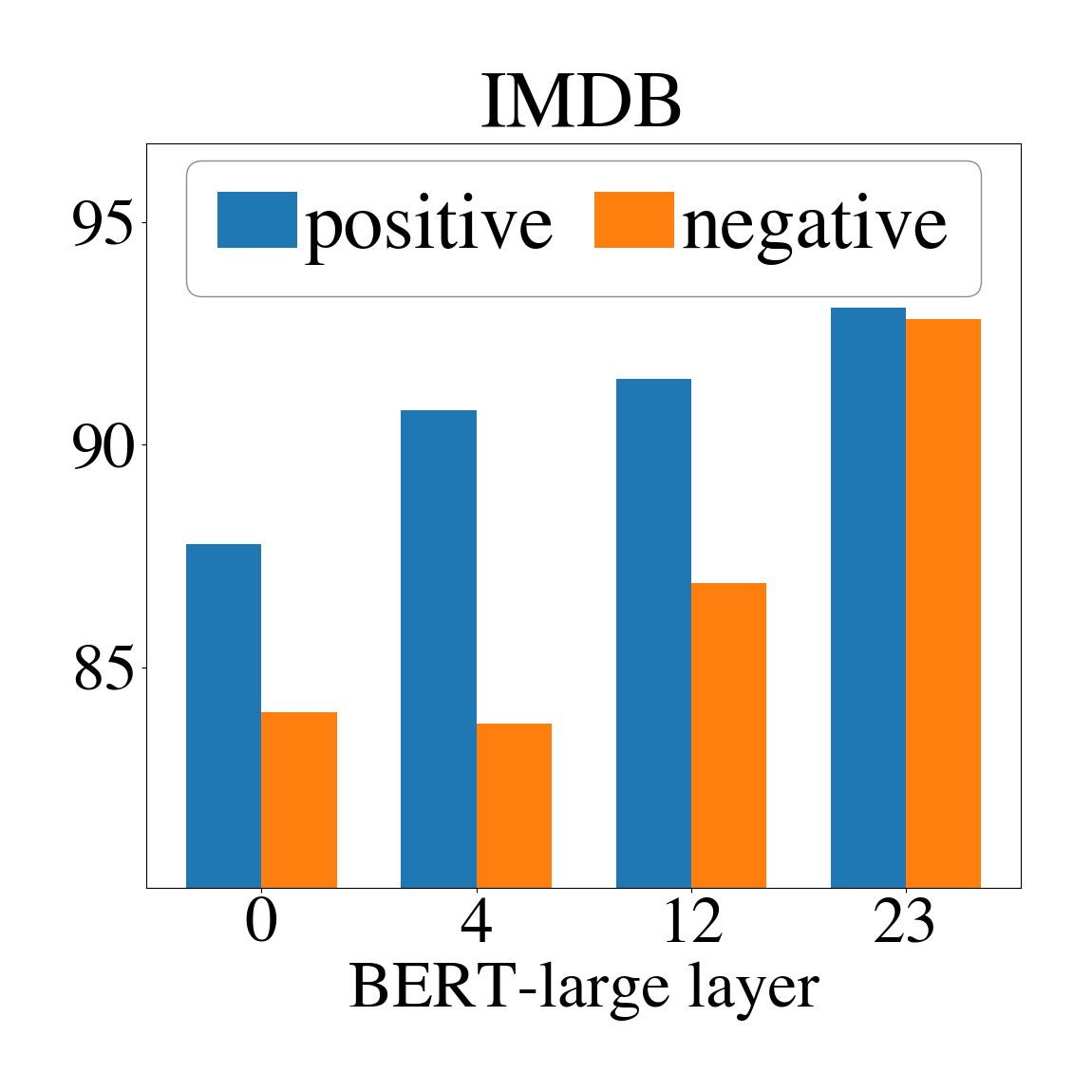}
    \includegraphics[trim={0cm 0 0cm 0}, clip,width=\figlen]{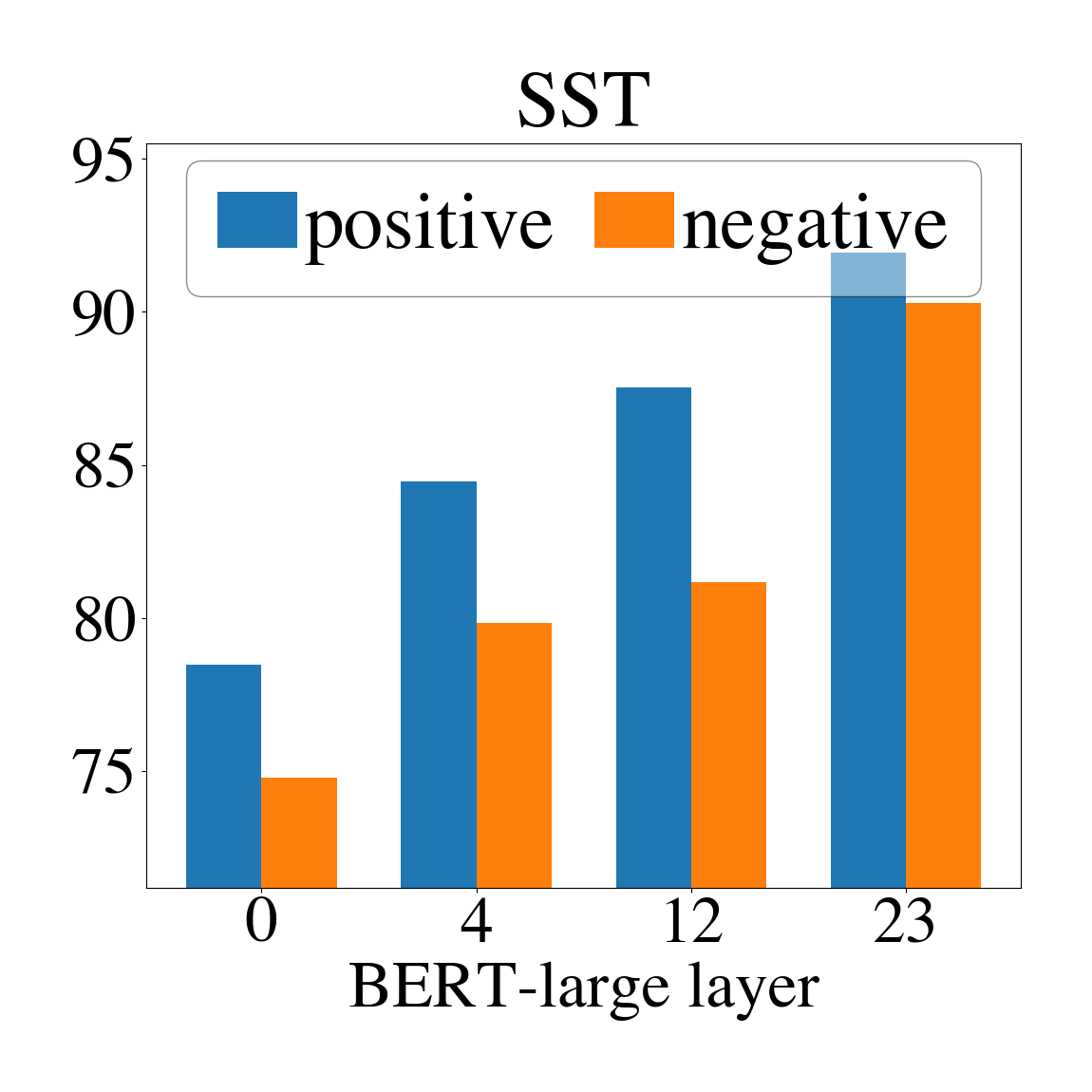}
    \includegraphics[trim={0cm 0 0cm 0cm}, clip,width=\figlen]{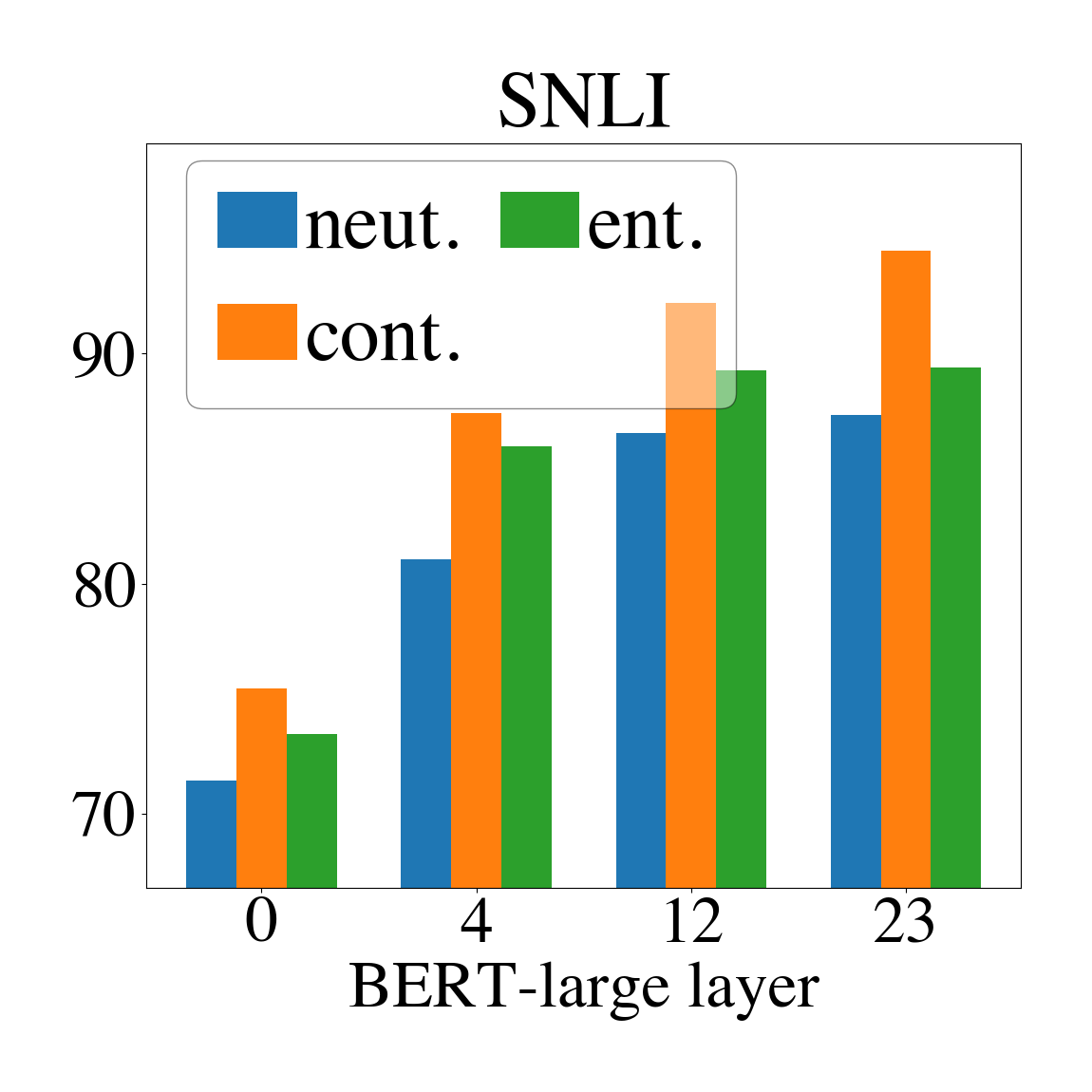}
    \includegraphics[trim={0cm 0 0cm 0cm}, clip,width=\figlen]{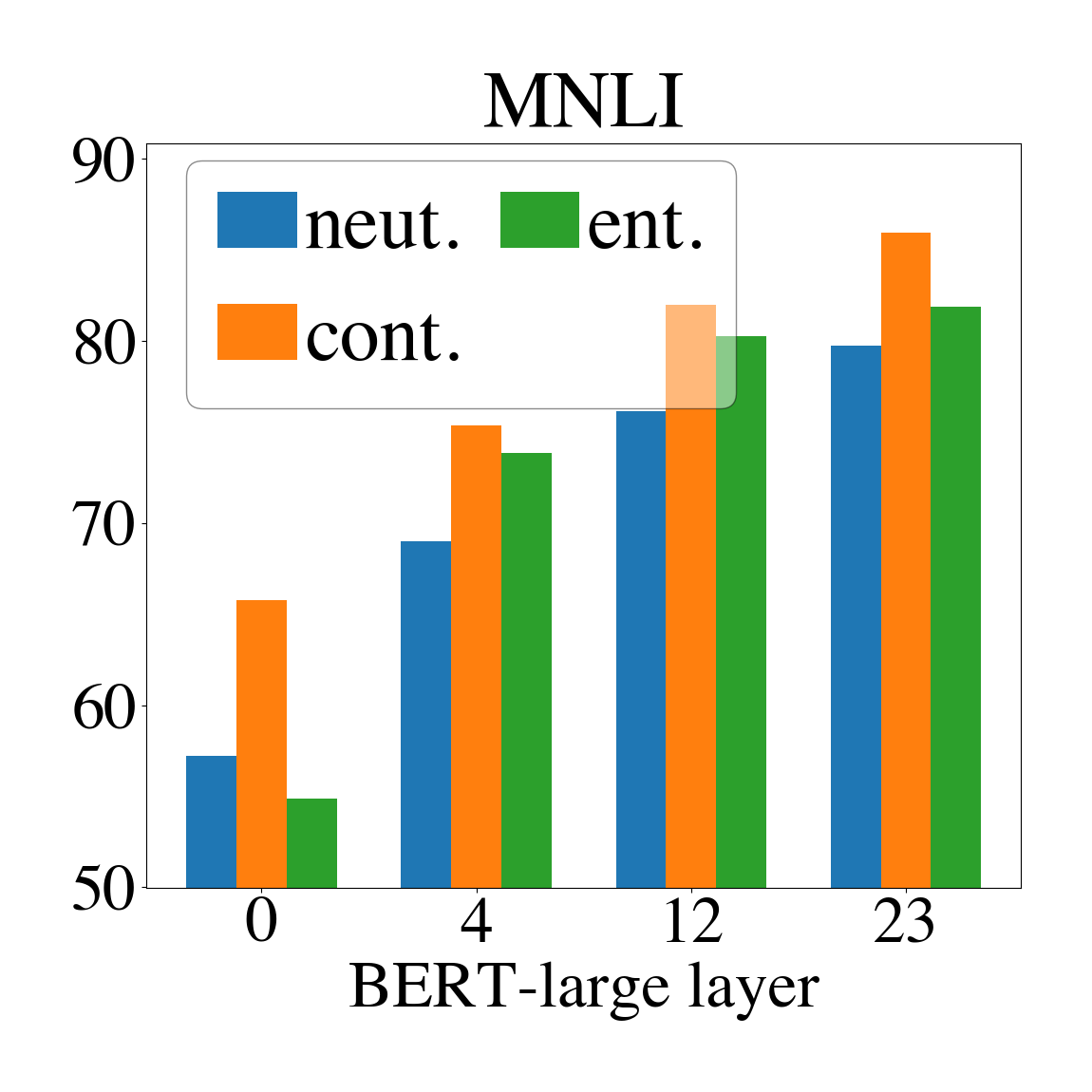}
    \caption{\label{fig:label_analysis} 
    Instances with different labels are predicted with different degrees of confidence.}
\end{figure*}

\begin{figure*}[t]
\newcommand{\figlen}[0]{.4\columnwidth}
    \centering
    \includegraphics[trim={0cm 0 0cm 0}, clip,width=\figlen]{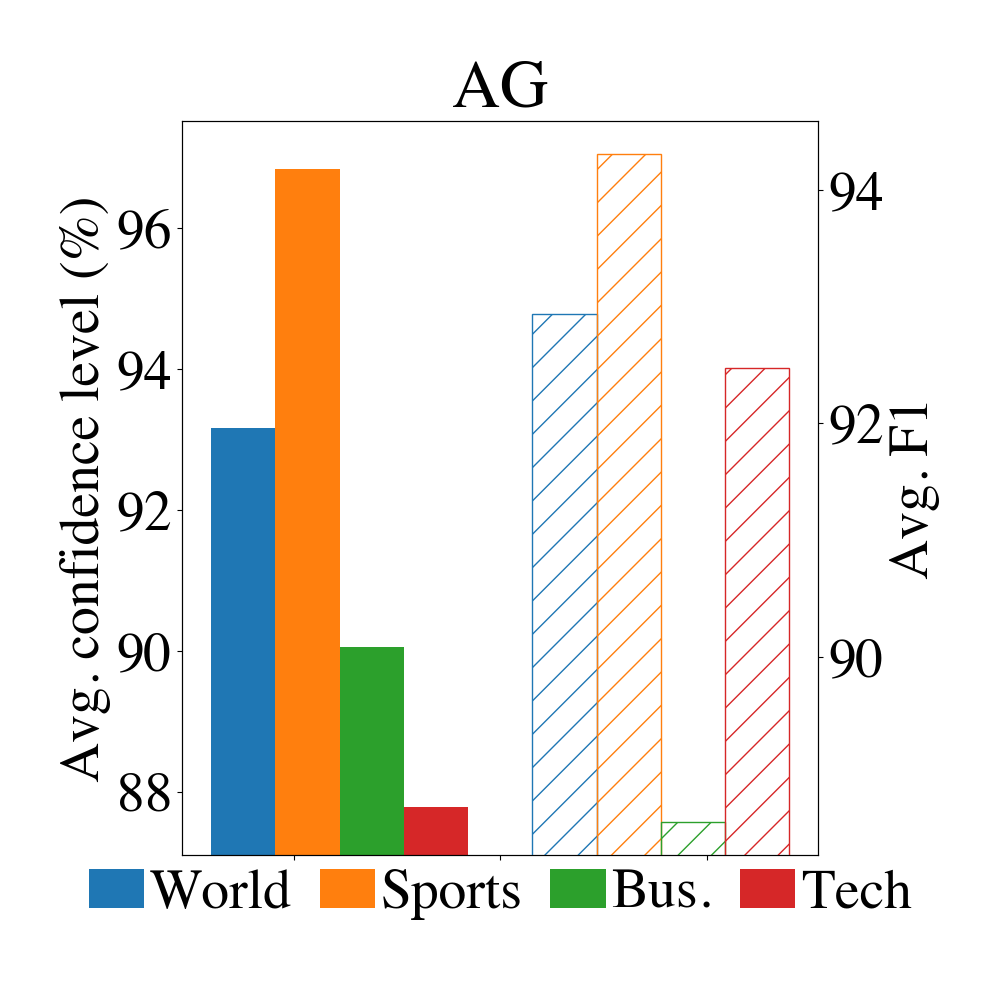}
    \includegraphics[trim={0cm 0 0cm 0}, clip,width=\figlen]{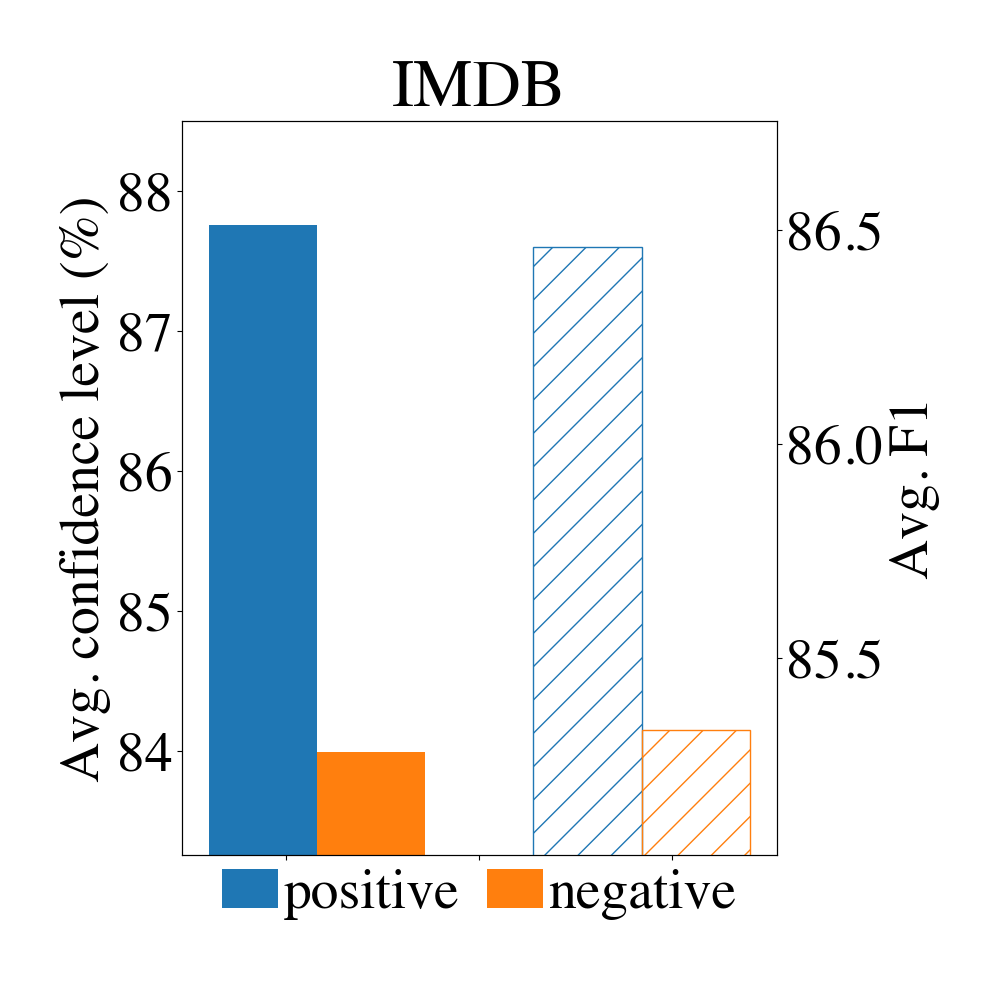}
    \includegraphics[trim={0cm 0 0cm 0}, clip,width=\figlen]{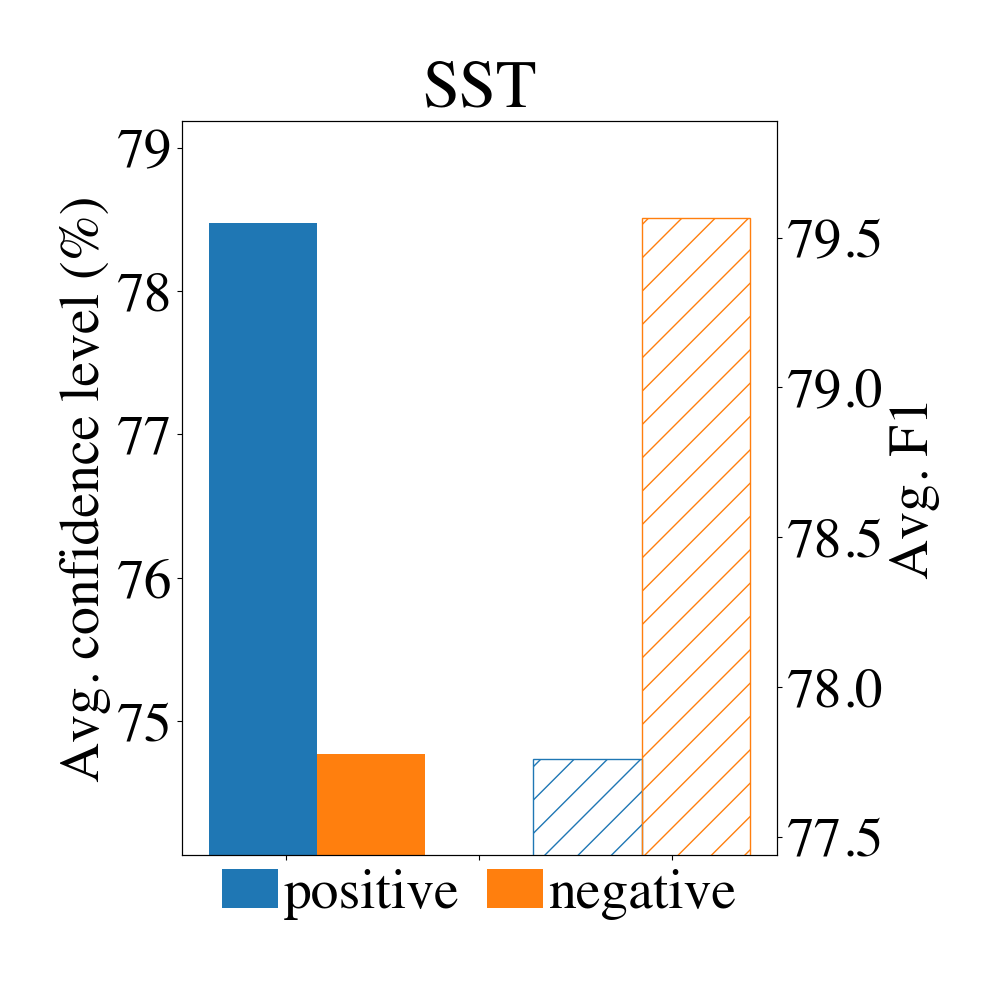}
    \includegraphics[trim={0cm 0 0cm 0cm}, clip,width=\figlen]{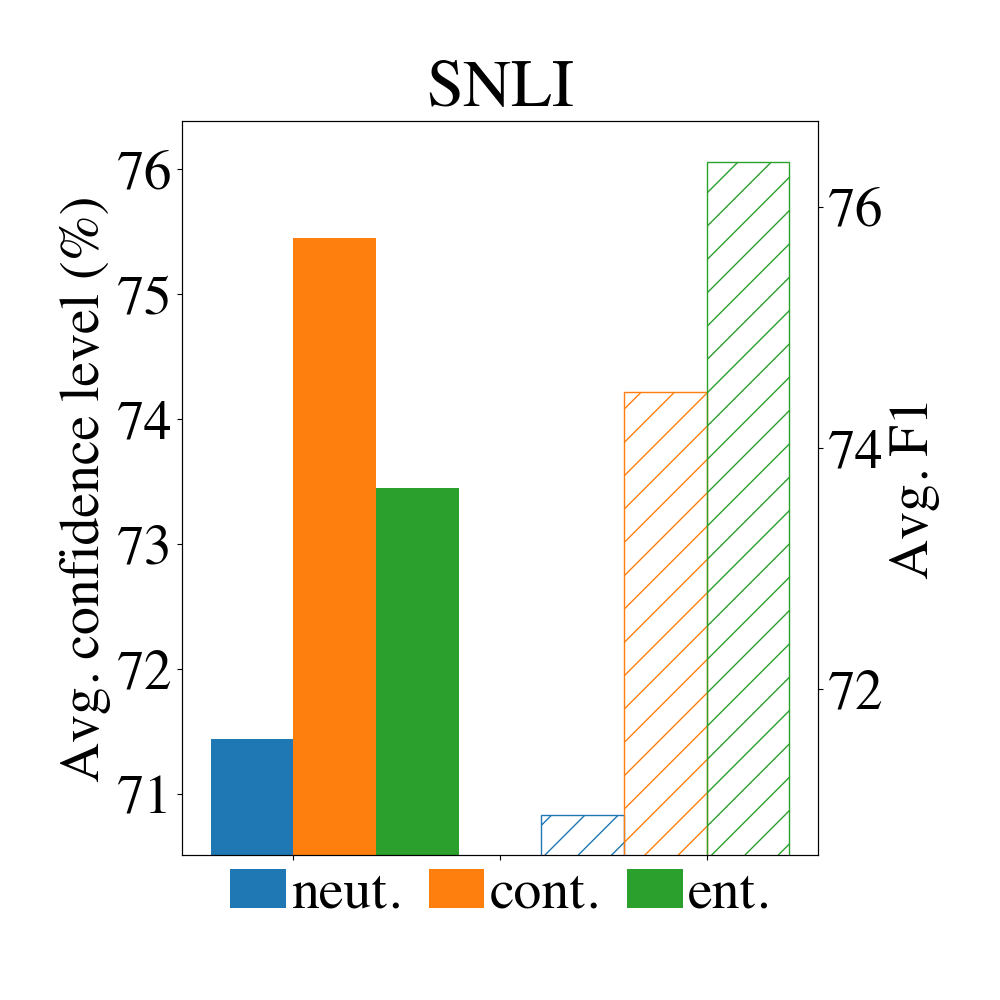}
    \includegraphics[trim={0cm 0 0cm 0cm}, clip,width=\figlen]{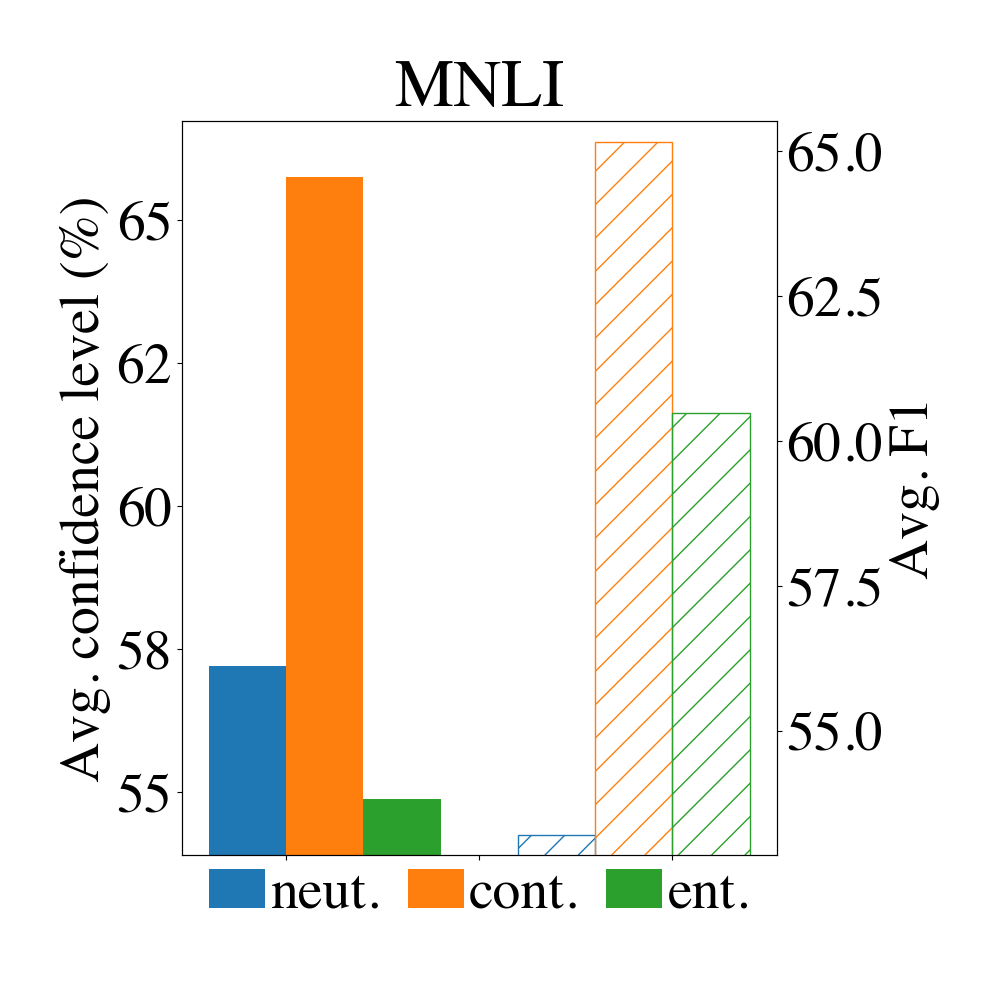}
    \caption{\label{fig:confidence_accuracy} 
    Comparing confidence levels and $F_1$ scores of our most efficient classifier across datasets and labels.
    High confidence by the model is sometimes explained by ``easy'' classes that are predicted with high $F_1$ (e.g., \textit{sports} in AG). Other cases might stem from biases of the model which make it overconfident despite the label being harder than other labels (e.g., \textit{positive} in SST).}
\end{figure*}

\paragraph{Confidence across labels} 
\figref{label_analysis} shows the proportion of instances in our validation set that are predicted with high confidence by our calibrated model (90\% threshold) for each dataset, label, and model size. 
We first note that across all datasets, and almost all model sizes, different labels are not predicted with the same level of confidence. 
For instance, for AG, the layer 0 model predicts the \emph{tech} label with 87.8\% average confidence, compared to 96.8\% for  the \emph{sports} label. 
Moreover, in accordance with the overall performance, across almost all datasets and model sizes, the confidence levels increase as the models get bigger in size.
Finally, in some cases, as we move towards  larger models, the gaps in confidence close (e.g., IMDB and SST), although the relative ordering hardly ever changes.

Two potential explanations come up when observing these results; either some labels are easier to predict than others (and thus the models are more confident when predicting them), or the models are biased towards some classes compared to others.
To help differentiate between these two hypotheses, we plot in \figref{confidence_accuracy} the average confidence level and the average $F_1$ score of the most efficient classifier across labels and datasets. 

The plot indicates that both hypotheses are correct to some degree.
Some labels, such as \textit{sports} for AG and \textit{positive} for IMDB, are both predicted with high confidence, and solved with high accuracy. 
In contrast, our model is overconfident in its prediction of some labels (\textit{business} for AG, \textit{positive} for SST), and underconfident in others (\textit{tech} for AG, \textit{entailment} for MNLI).
These findings might indicate that while our method is designed to be globally calibrated, it is not necessarily calibrated for each label individually.  Such observations relate to existing concerns regarding fairness when using calibrated classifiers \cite{Pleiss:2017}.

\section{Related Work}\label{sec:related}

Methods for making inference more efficient have received considerable
attention in NLP over the years \cite[][\textit{inter alia}]{Eisner:1999,Goldberg:2010}.
As the field has converged on deep neural architecture solutions, most efforts focus on making models smaller (in terms of
model parameters) in order to save space as well as potentially speed
up inference.

In \textit{model distillation} \cite{Hinton:2014} a smaller model (the student) is trained to mimic the behavior or structure of the original, larger model (the teacher).
The result is typically a student that is as accurate as the teacher, but smaller and faster \cite{Kim:2016a,Jiao:2019,Tang:2019,Sanh:2019}.
\textit{Pruning} \cite{Lecun:1990} removes some of the weights in the network, resulting in a smaller, potentially faster network.
The basic pruning approach removes individual weights from the network \cite{Swayamdipta:2018,Gale:2019}.
More sophisticated approaches induce structured sparsity, which removes full blocks  \cite{Michel:2019,Voita:2019,Dodge:2019b}.
\citet{Liu:2018} and \citet{Fan:2020} pruned deep models by applying dropout to different layers, which allows dynamic control of the speed/accuracy tradeoff of the model without retraining. 
Our method also allows for controlling this tradeoff with a single training pass, and yields computational savings in an orthogonal manner: by making early exit decisions. 

\emph{Quantization} is another popular method to decrease model size, 
which reduces the numerical precision of the model's weights, and therefore both speeds up numerical operations and reduces model size \cite{Wrobel:2018,Shen:2019,Zafrir:2019}.

Some works introduced methods to allocate fewer resources to certain parts of the input (e.g., certain words), thereby potentially reducing training and/or inference time \cite{Graves:2016,Seo:2018}.
Our method also puts less resources into some of the input, but does so at the document level rather than for individual tokens.

A few concurrent works have explored similar ideas for dynamic early exits in the transformer model. 
\citet{Elbayad:2020} and \citet{Dabre:2020} introduced early stopping for sequence-to-sequence tasks (e.g., machine translation).
\citet{Bapna:2020} modify the transformer architecture with ``control symbols'' which determine whether components are short-circuited to optimize budget. 
Finally, \citet{Liu:2020} investigated several inference-time cost optimizations (including early stopping) in a multilingual setting.

Several computer vision works explored similar ideas to the one in this paper. 
\citet{Wang:2018} introduced a method for dynamically skipping convolutional layers. 
\citet{Bolukbasi:2017} and \citet{Huang:2018} learned early exit policies for computer vision architectures, observing substantial  computational gains. 

\section{Conclusion}
We presented a method that improves the speed/accuracy tradeoff for inference using pretrained language models.
Our method makes early exits for simple instances that require less processing, and thereby avoids running many of the layers of the model.
Experiments with BERT-large on five text classification and NLI datasets yield substantially faster inference compared to the standard approach, up to 80\% faster while maintaining similar performance.
Our approach requires neither additional training time nor significant number of additional parameters compared to the standard approach. It also allows for controlling the speed/accuracy tradeoff using a single model, without retraining it for any point along the curve.
\section*{Acknowledgments}
The authors thank the members of Noah's ARK at the University of Washington, the researchers at the Allen Institute for AI, and the anonymous reviewers for their valuable feedback.

\clearpage

\bibliography{acl2020}
\bibliographystyle{acl_natbib}

\newpage
\appendix

\section{Implementation Details}\label{app:implementation}
We fine-tune both our model and our baselines with dropout 0.1. 
We run all our experiments on a single Quadro RTX 8000 GPU.
Our model is implement using the AllenNLP library \cite{Gardner:2018}.\footnote{\url{https://allennlp.org}}
Our calibration code relies on the implementation of \citet{Guo:2017}.\footnote{\url{https://github.com/gpleiss/temperature_scaling}}

We fine-tune text classification models for 2 epochs and NLI models for 4 epochs.
We run ten trials of random search on the validation set for both our model and our baselines to select both a learning rate among \{0.00002, 0.00003, 0.00005\} and a random seed.
For our baselines, we select the highest performing model on the validation set among the ten runs. 
For our model, we select the
one with the highest performance  averaged across all thresholds explored (we use 0\% and 5\% intervals in the range [55\%, 100\%]) on the validation set.

\section{Validation Results}\label{app:dev_results}

\begin{figure}[!h]
\newcommand{\figlen}[0]{.8\columnwidth}
    \centering
    \includegraphics[trim={0cm 0 0cm .5cm}, clip,width=\figlen]{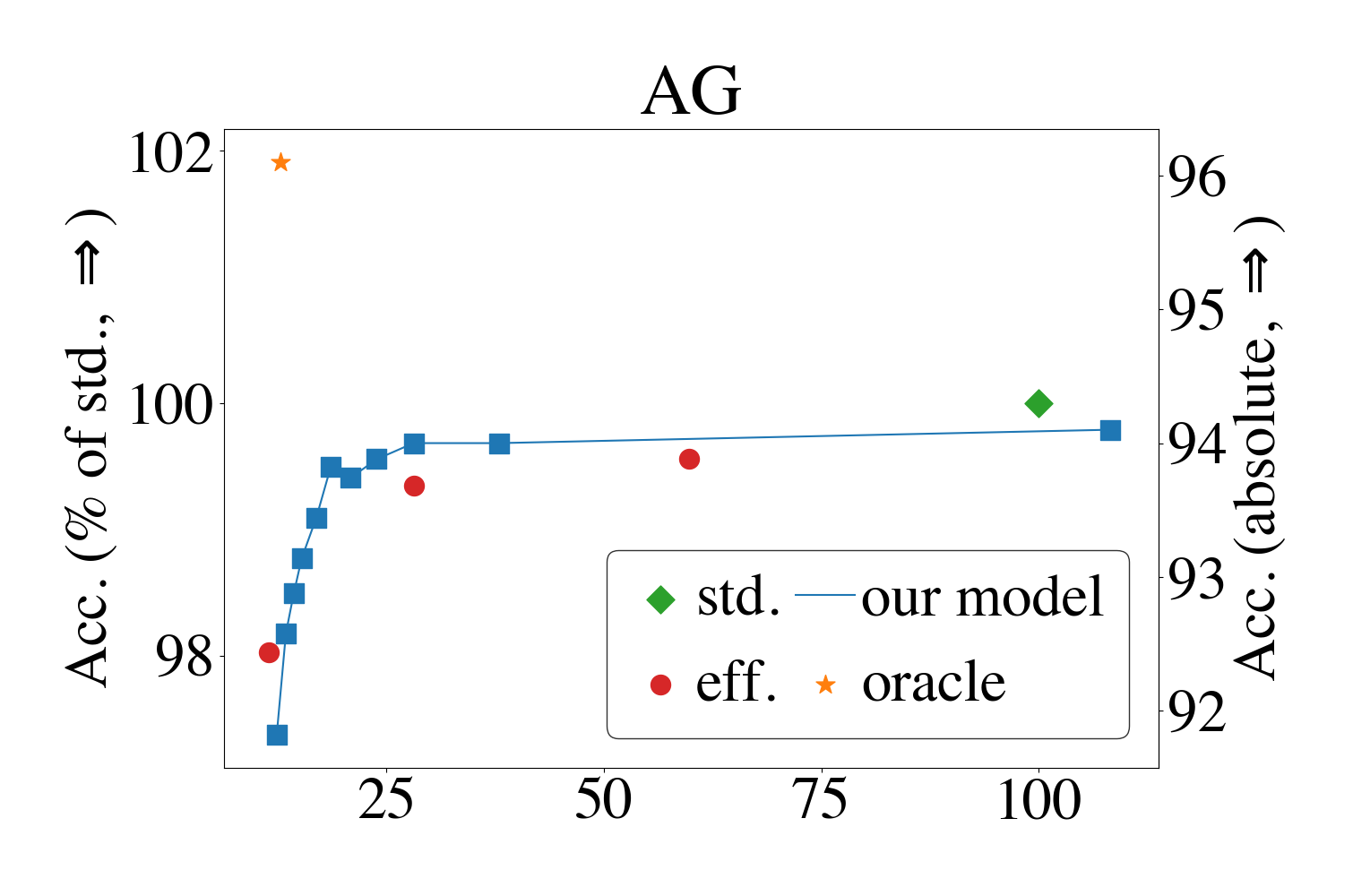}
    \includegraphics[trim={0cm 0 0cm .5}, clip,width=\figlen]{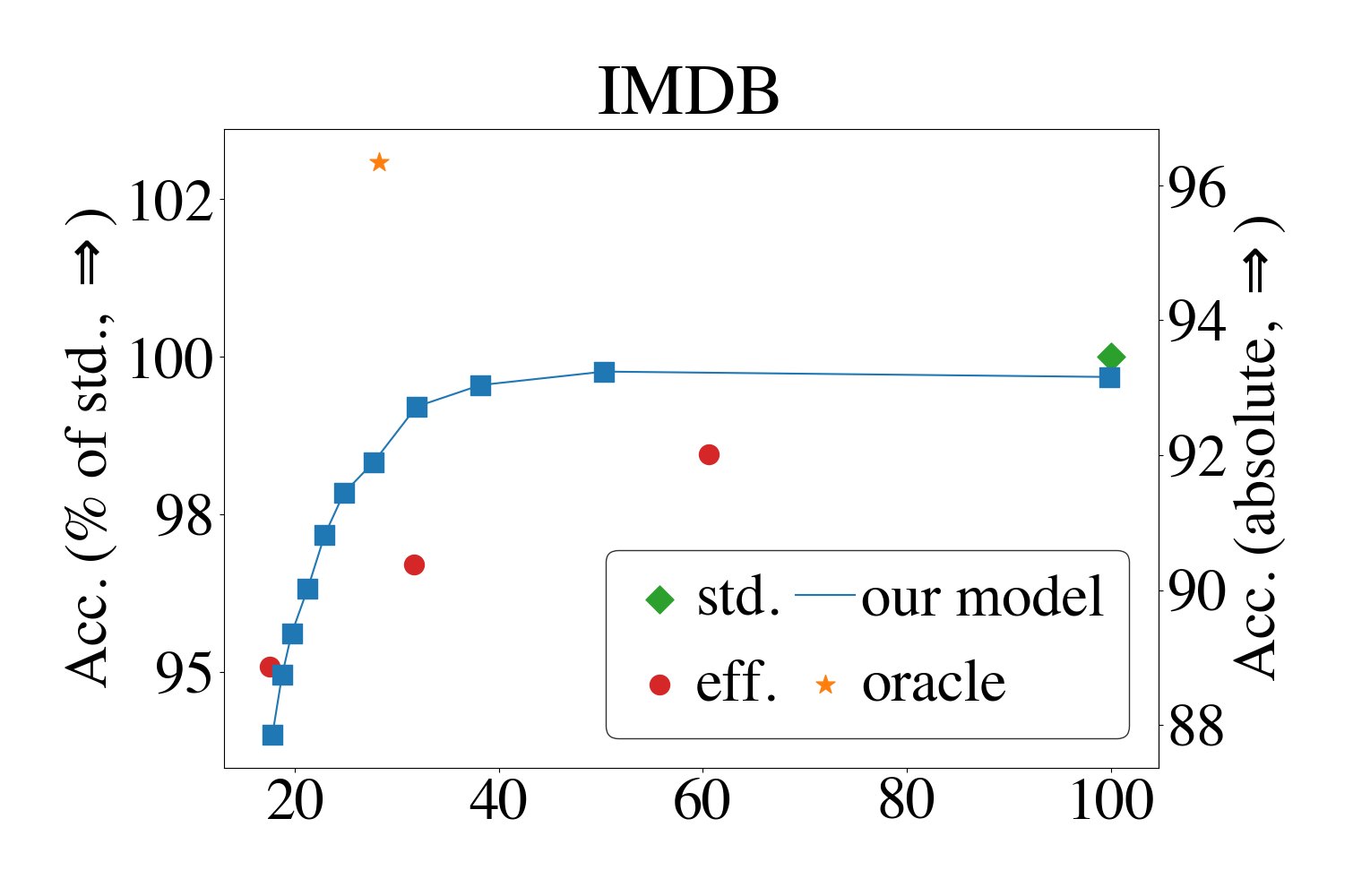}
    \includegraphics[trim={0cm 0 0cm .5cm}, clip,width=\figlen]{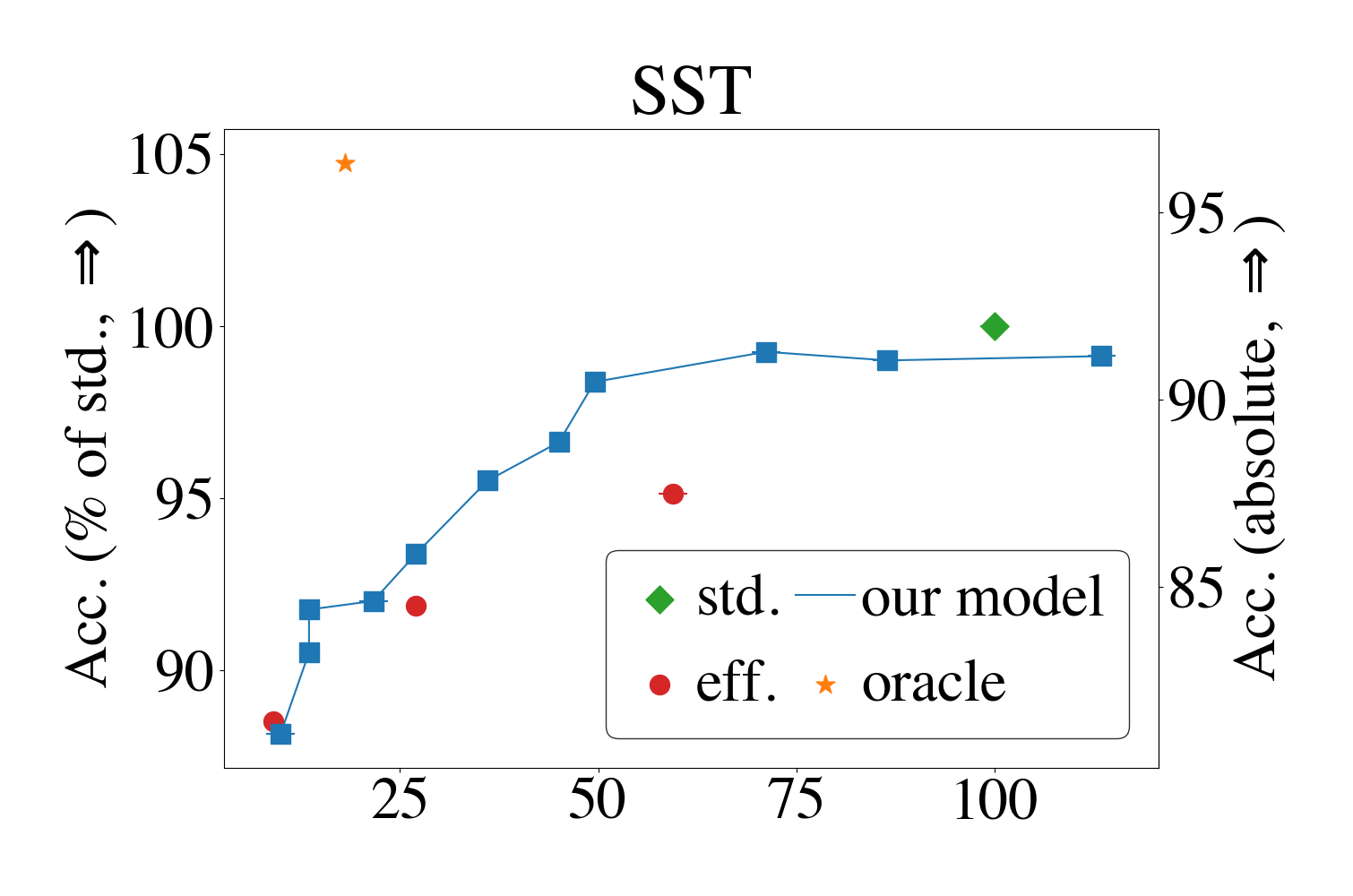}
    \includegraphics[trim={0cm 0 0cm .5cm}, clip,width=\figlen]{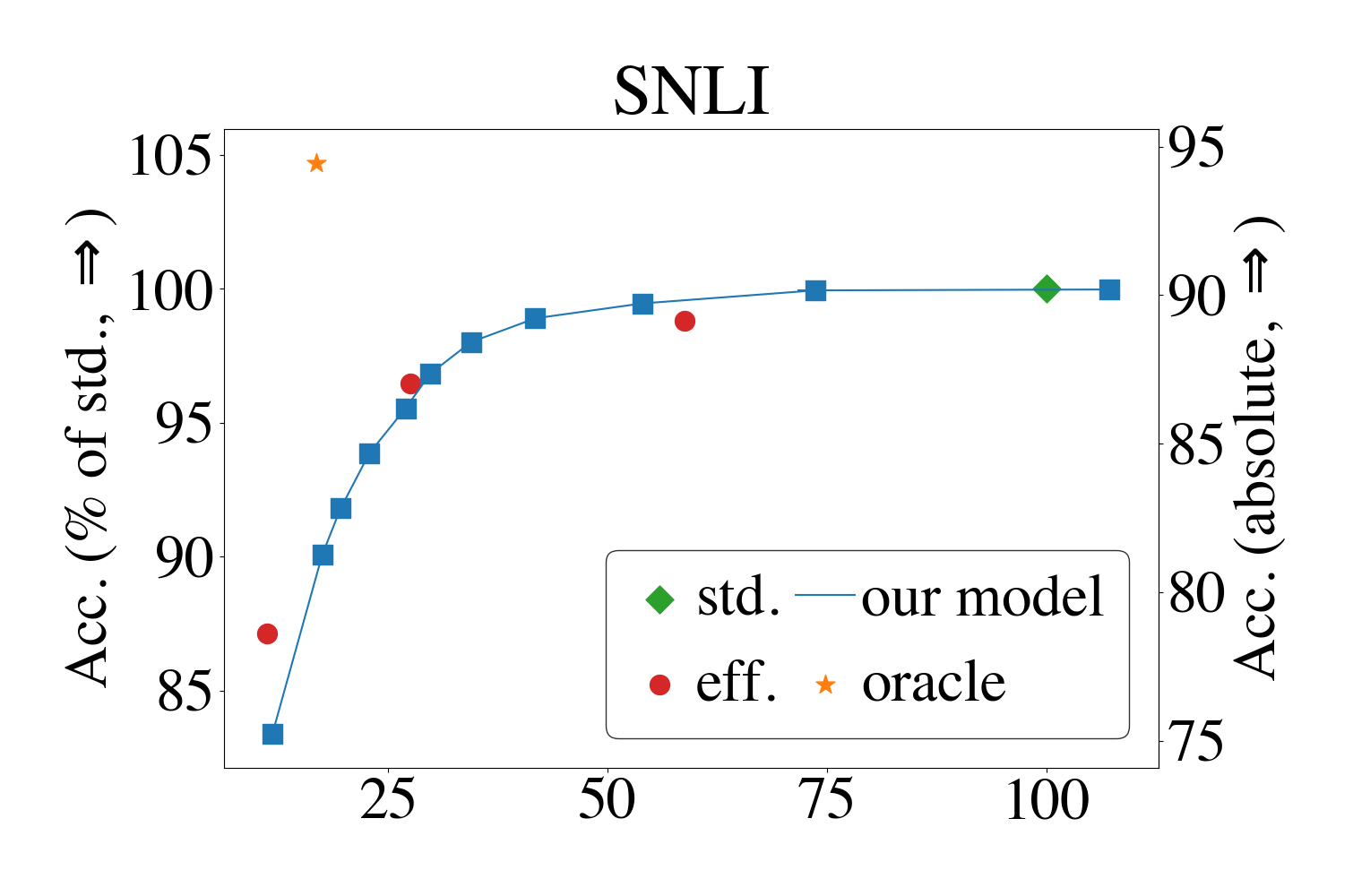}
    \includegraphics[trim={0cm 0 0cm .5cm}, clip,width=\figlen]{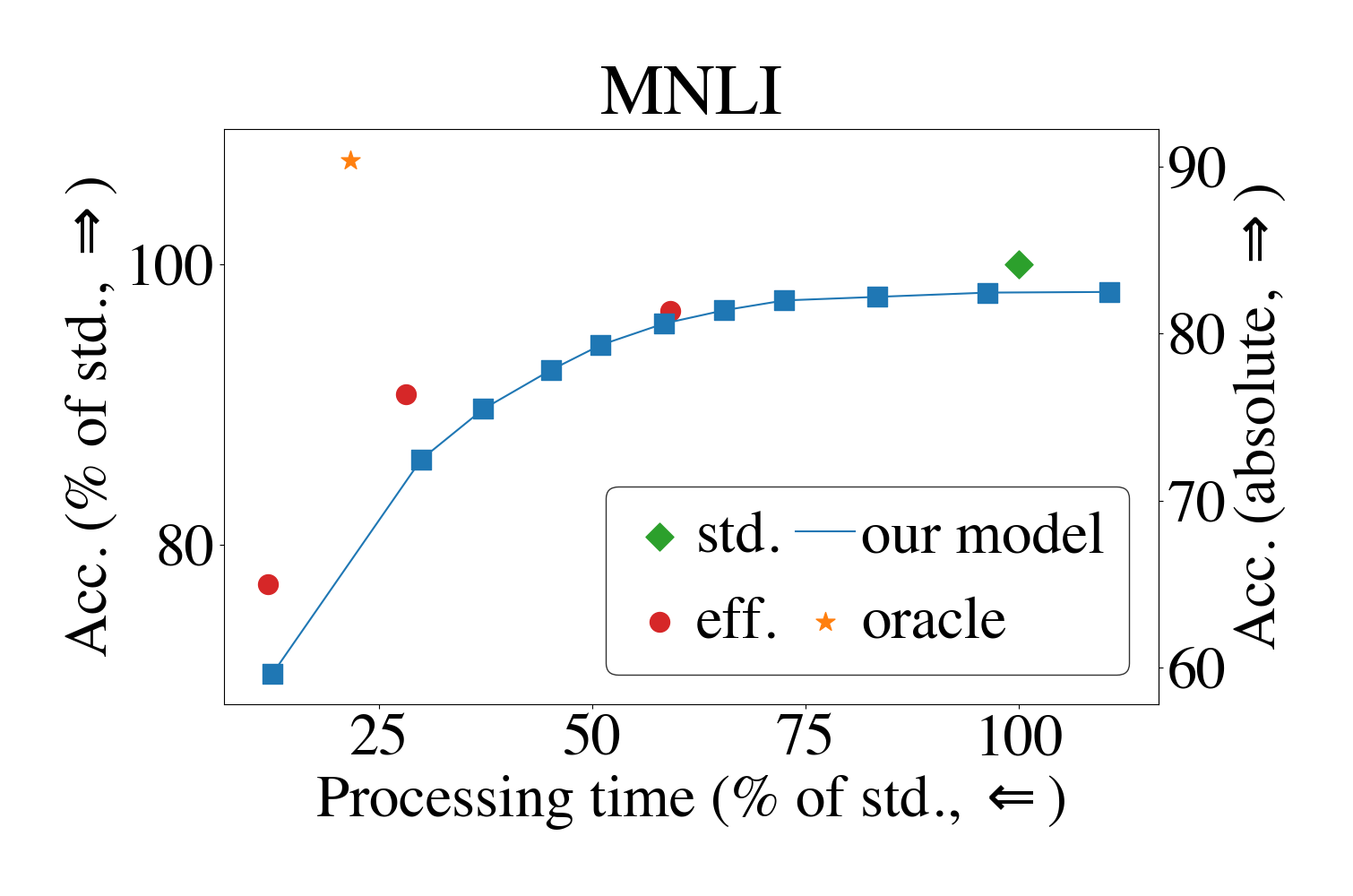}
    \caption{\label{fig:val_results} 
    Validation accuracy and processing time of our approach ({\color{blue}{blue line}}) and our standard baseline (std., {\color{DarkGreen}{green diamond}}), our efficient baselines (eff., {\color{red}{red dots}}) and our oracle ({\color{orange}{orange star}}). Left and higher is better.}
\end{figure}

\figref{val_results} shows the validation results of our experiments.

\end{document}